\documentclass[sn-basic]{sn-jnl}

\usepackage{graphicx}%
\usepackage{multirow}%
\usepackage{amsmath,amssymb,amsfonts}%
\usepackage{amsthm}%
\usepackage{mathrsfs}%
\usepackage[title]{appendix}%
\usepackage{xcolor}%
\usepackage{textcomp}%
\usepackage{manyfoot}%
\usepackage{booktabs}%
\usepackage{algorithm}%
\usepackage{algorithmicx}%
\usepackage{algpseudocode}%
\usepackage{subcaption}%
\usepackage{graphicx}
\usepackage{submission}
\usepackage{macros}
\usepackage{listings}%

\setmode{final}

\newtheorem{example}{Example}%

\begin{document}

\title[Evidential uncertainty sampling for active learning]{\centering Evidential uncertainty sampling for active learning}

\author*[1]{\fnm{Arthur} \sur{Hoarau}}\email{arthur.hoarau@univ-rennes.fr}
\author[2]{\fnm{Vincent} \sur{Lemaire}}\email{vincent.lemaire@orange.com}
\author[1]{\fnm{Yolande} \sur{Le Gall}}\email{yolande.le-gall@univ-rennes.fr}
\author[1]{\fnm{Jean-Christophe} \sur{Dubois}}\email{jean-christophe.dubois@univ-rennes.fr}
\author[1]{\fnm{Arnaud} \sur{Martin}}\email{arnaud.martin@univ-rennes.fr}

\affil*[1]{\orgdiv{University of Rennes}, \orgname{CNRS, IRISA, DRUID}, \orgaddress{\city{Rennes}, \postcode{35000}, \country{France}}}
\affil[2]{\orgname{Orange innovation}, \city{Lannion}, \postcode{22300}, \country{France}}

\abstract{Recent studies in active learning, particularly in uncertainty sampling, have focused on the decomposition of model uncertainty into reducible and irreducible uncertainties. In this paper, the aim is to simplify the computational process while eliminating the dependence on observations. Crucially, the inherent uncertainty in the labels is considered, \emph{i.e.} the uncertainty of the oracles. Two strategies are proposed, sampling by Klir uncertainty, which tackles the exploration-exploitation dilemma, and sampling by evidential epistemic uncertainty, which extends the concept of reducible uncertainty within the evidential framework, both using the theory of belief functions. Experimental results in active learning demonstrate that our proposed method can outperform uncertainty sampling.}


\maketitle

\section{Introduction}\label{section:introduction}

\emph{Active Learning (AL) -} For reasons of efficiency, cost or energy reduction in machine learning or deep learning, one of the important issues is related to the amount of data and in some cases, to the amount of labeled data. Active learning~\citep{Settles2009} is a part of machine learning in which the learner can choose which observation to label in order to work with only a fraction of the labeled dataset to reduce the labeling cost. While primarily used for cost reduction~\citep{Hacohen2022}, active learning finds application in various domains like anomaly detection, as seen in~\citep{Naoki2006} and~\citep{Timo2023}.
The essence of active learning lies in empowering the learner to strategically label selected observations. Among all the proposed strategies in the literature~\citep{Settles2009,Aggarwal2014}, one of the most recognized is uncertainty sampling~\citep{Lewis1994,Nguyen2022}.\\\\
\emph{Uncertainty Quantification (UQ) -} finds applications across various fields, including medical image analysis, as discussed in a review by~\citep{huang2023review}, and in-depth exploration of deep learning applications and techniques~\citep{ABDAR2021243}, such as recent evidential deep learning~\citep{Sensoy2018}. Concerning frameworks for uncertainty, numerous methods exist for quantifying uncertainty, with many applied to credal sets as reviewed by~\citep{hullermeier22}, or evidential entropies ~\citep{Deng2020}. Despite being described several years ago~\citep{Hora1996}, recent literature~\citep{eyke2019,kendall2017,senge2014,Charpentier2020} distinguishes two main types of uncertainty: epistemic and aleatoric. Aleatoric uncertainty arises from the stochastic property of the event and is therefore not reducible, whereas epistemic uncertainty is related to a lack of knowledge and can be reduced. Most proposed calculations hinge not only on the model predictions but also on parameter estimations derived directly from the observations themselves.\\\\
\emph{(AL) $\cup$ (UQ) - What is the issue? -} In uncertainty sampling, the learner selects the instances for which it is most uncertain. Until recently, the literature has mostly proposed measures to quantify this uncertainty, such as entropy, in a probabilistic form. But this kind of uncertainty cannot exploit and capture the difference between a label given by someone who has hesitated for a long time and a label given by someone who has no doubt, and therefore uncertainty that may already exist in the labels. In this paper, we propose to use evidential reasoning within the context of active learning. Moreover we propose eliminating direct dependence on the observations and advocating for solely utilizing the model output to achieve a comparable decomposition (\emph{i.e} epistemic-aleatoric) of uncertainty. This also tackles the exploration-exploitation issue in active learning, with the possibility of choosing one or the other, or even a compromise as suggested by~\citep{Bondu2010}. This paper will use the theory of belief functions that generalises probabilities and will use it in the context of active learning.\\\\
\begin{figure}[h!]
    \centering
    \begin{subfigure}{\linewidth}
    \centering
    \includegraphics[width=0.9\linewidth]{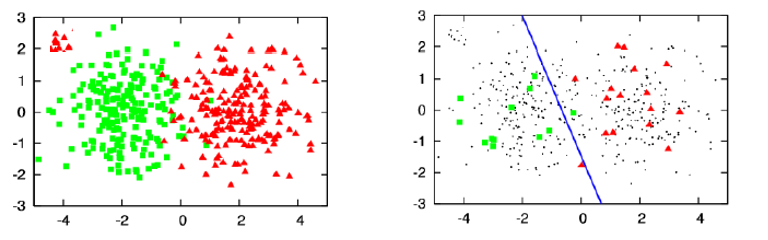}
    \end{subfigure}
    \caption{Illustrating the exploration-exploitation dilemma in active learning: complete dataset vs. active learning iterations.}\label{fig:dilemma}
\end{figure}
\emph{Note on Exploration-Exploitation Dilemma -} In Figure~\ref{fig:dilemma}, we present a visual depiction of the exploration-exploitation dilemma. In a 2D classification task, the left panel portrays a scenario where all observations are labeled, while the right panel reflects the outcome after iterative labeling rounds, resulting in sparse observations.
Should the sampling strategy heavily favor exploitation, as denoted by the blue line representing the classifier, it is evident that the classifier will neglect further investigation into the top left corner. Consequently, upon encountering the red examples in this unexplored territory later in the process, the classifier's performance undergoes a significant decline, necessitating extensive parameter adjustments. Conversely, a purely exploratory strategy prolongs the duration required to unveil critical patterns, including the ``two red patterns'' mentioned. Thus, an optimal strategy must delicately balance exploration and exploitation to navigate this trade-off effectively.\\\\
\emph{Contributions of this paper -} The primary goal of this paper is to consider the uncertainty inherent in the labels (introduced by the entities labeling observations, referred to as "Oracles" in active learning), and to address the exploration-exploitation dilemma, during sampling. To this end we propose two uncertainty sampling strategies capable of representing a decomposition of the model uncertainties with regard to the uncertainty already present in the labels: (i) a first strategy which is based upon two different uncertainties, discord - how self-conflicting the information is - and non-specificity - how imprecise the information is - in the model output; and (ii) a second strategy which extends the epistemic uncertainty to the evidential framework and to several classes, thus simplifying the computation.  To succeed in this challenge we use evidential models able to handle such uncertain labels, such as~\citep{denoeux1995,ELOUEDI2001,Denoeux2000,Yuan2020}. By doing this, one can effectively distinguish and account for the difference between a label provided by someone who has hesitated extensively and a label given by someone who has no doubts. As a result, we can identify and quantify the uncertainty inherent in the labels themselves.\\\\
\emph{Organization  -} The paper is organized as follows: section~\ref{section:preliminaries} introduces some important notions of imperfect labeling and the modeling of these richer labels using the theory of belief functions. The conventional uncertainty sampling approach is also recalled and section~\ref{section:epistemic} describes the separation between aleatoric and epistemic uncertainties. Section~\ref{section:method} introduces the two new proposed strategies and section~\ref{section:experiments} presents the experiments\footnote{For details on experiments conducted in theoretical sections, visit: \url{https://anonymous.4open.science/r/evidential-uncertainty-sampling-D453}}, first on a real world dataset with rich labels and then in active learning to highlight the relevance and efficacy of the proposed method. Section~\ref{section:discussion} discusses the encountered limits and section~\ref{section:conclusion} concludes the article.

\section{Preliminaries}\label{section:preliminaries}

In this section, we provide foundational knowledge essential for understanding the rest of the paper, beginning with rich labels, which are characterized by the theory of belief functions, and concluding with the classical method of uncertainty sampling.

\subsection{Imperfect labeling}\label{section:rich_labels}
Most of the datasets used for classification consider hard labels, with a binary membership where the observation is either a member of the class or not. In this paper, we refer as rich labels the elements of response provided by a source that may include several degrees of imprecision (\emph{i.e.} ``\emph{This might be a cat}'', ``\emph{I don't know}'' or ``\emph{I am hesitating between dog and cat, with a slight preference for cat})''. Such datasets, offering uncertainty already present in the labels, exist~\citep{Thierry2022} but are not numerous. These labels are called rich in this paper since they provide more information than hard labels and can be modeled using the theory of belief functions.

\subsection{Theory of belief functions}

\label{section:dempster} The theory of belief functions introduced by~\citep{Dempster1967} and~\citep{shafer1976}, is used in this study to model uncertainty and imprecision for labeling and prediction. 
Let $\Omega = \{\omega_1, \ldots, \omega_M\}$ be the frame of discernment for $M$ exclusive and exhaustive hypotheses. In supervised learning, this refers to the labels (i.e., classes), or the output space. It is assumed that only one element of $\Omega$ is true (closed-world assumption~\citep{Smets1994}).
The power set $2^\Omega$ is the set of all subsets of $\Omega$.
A mass function assigns the belief that a source may have about the elements of the power set of $\Omega$, such that the sum of all masses is equal to $1$.
\begin{equation} \label{eq:unity}
    m: 2^\Omega\rightarrow [0, 1],    \sum_{A\in 2^\Omega} m(A) = 1.
\end{equation}

Each subset $A\in 2^\Omega$ such as $m(A) > 0$ is called a \emph{focal element} of $m$. The uncertainty is therefore represented by a mass $m(A) < 1$ on a focal element $A$ and the imprecision is represented by a non-null mass $m(A) > 0$ on a focal element $A$ such that $|A| > 1$.

A mass function $m$ is called \emph{categorical mass function} when it has only one focal element such that $m(A) = 1$. In the case where $A$ is a set of several elements, the knowledge is certain but imprecise. For $|A| = 1$, the knowledge is certain and precise.

On decision level, the pignistic probability $BetP$ of~\citep{Smets1994} helps decision making on singletons:
\begin{equation}\label{eq:betp}
    BetP(\omega) = \sum_{A\in 2^\Omega, \;\omega\in A}\frac{m(A)}{|A|}.
\end{equation}

It is also possible to combine several mass functions (beliefs from different sources) into a single body of evidence. 
If the labels and therefore the masses are not independent, a simple average of the mass functions $m_j$ derived from $N$ sources can be defined as follows:
\begin{equation} \label{eq:mean}
    m(A) = \displaystyle\frac{1}{N}\sum_{j = 1}^{N}{m_j(A)}, \;\;A \in 2^\Omega. \\
\end{equation}
There are other possible combinations that are more common than the mean, many of which are listed by~\citep{Martin2019}.

\begin{example}
Let $\Omega = \{Cat, Dog\}$ be a frame of discernment. An observation labeled ``Cat'' by a source can be modeled in the framework of belief functions by the mass function $m_1$ such as: $m_1(\{Cat\}) = 1$ and $m_1(A) = 0, \;\forall A \in 2^\Omega\backslash\{Cat\}$.
\end{example}

\begin{example}
An observation labeled ``Cat or Dog'' by a source can be modeled by the mass function $m_2$ such as: $m_2(\{Cat, Dog\}) = 1$ and $m_2(A) = 0$, $\forall A \in 2^\Omega\backslash\{Cat, Dog\}$.
\end{example}

\begin{example}
The average mass function $\bar{m}$ of $m_1$ and $m_2$ is: $\bar{m}(\{Cat\}) = 0.5$, $\bar{m}(\{Cat, Dog\}) = 0.5$ and $\bar{m}(A) = 0$ for all other subsets $A$ in $2^\Omega$. 
Its pignistic probability $BetP$, used for decision making, is given as follows: $BetP(\{Cat\}) = 0.75$ and $BetP(\{Dog\}) = 0.25$.
\end{example}

\subsection{Uncertainty sampling}\label{subsection:uncertainty}Active learning iteratively builds a training set by selecting the best instances to label. The principle is to label as few observations as possible for a given performance or to achieve the best possible performance within a given budget. Among all the strategies proposed in the literature~\citep{Settles2009} one of the best known methods is uncertainty sampling~\citep{Lewis1994}, where the function that defines the instances to be labeled maximizes the uncertainty related to the model prediction as described below.

Let $\mathcal{U}$ be the uncertainty to label a new observation $x$ for a given model and $\Omega = \{\omega_1, \ldots, \omega_M\}$ the set of the $M$ possible classes. The uncertainty $\mathcal{U}$ can be calculated in several ways, a classical approach is to use Shannon's entropy:
\begin{equation}\label{eq:entropy}
    {\mathcal{U}}(x) = - \sum_{\omega\in\Omega}p(\omega|x)\text{log}[p(\omega|x)],
\end{equation}
with $p(\omega|x)$ the probability for $x$ to belong to the class $\omega$, given by the model. Other common uncertainty criteria include the least confidence measure:
\begin{equation}\label{eq:least_confidence}
    {\mathcal{U}}(x) = 1 - \underset{\omega\in\Omega}{\text{max}}[p(\omega|x)].
\end{equation}
Measuring the uncertainty of a model to predict the class of some observations can be useful to identify the areas of uncertainty in a space.

\begin{figure}
    \centering
    \begin{subfigure}{.25\linewidth}
    \includegraphics[width=\linewidth]{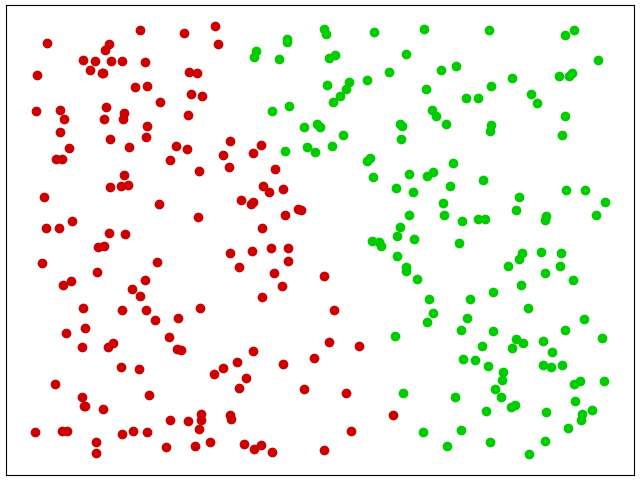}
    \end{subfigure}
    \begin{subfigure}{.25\linewidth}
    \includegraphics[width=\linewidth]{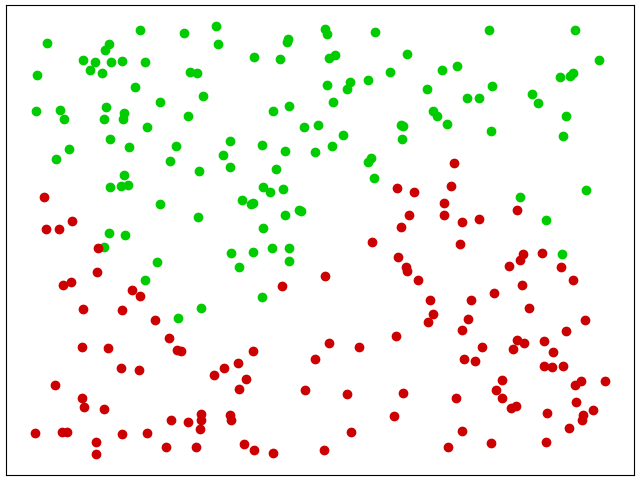}
    \end{subfigure}
    \begin{subfigure}{.25\linewidth}
    \includegraphics[width=\linewidth]{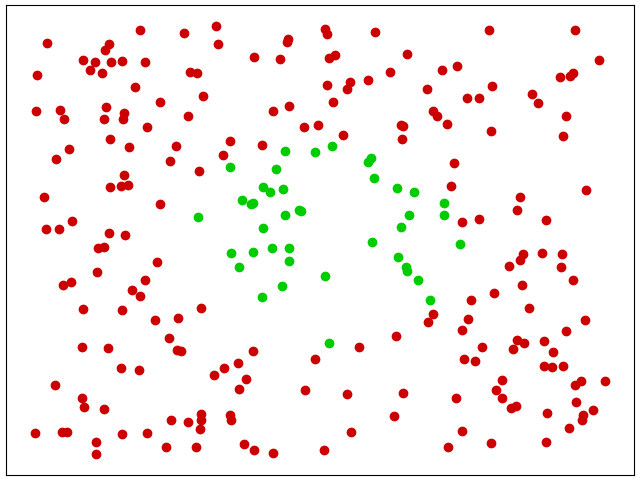}
    \end{subfigure}\\
    \begin{subfigure}{.25\linewidth}
    \includegraphics[width=\linewidth]{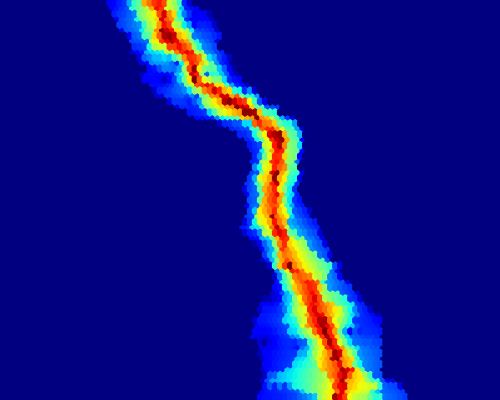}
    \caption{}
    \label{subfig:line}
    \end{subfigure}
    \begin{subfigure}{.25\linewidth}
    \includegraphics[width=\linewidth]{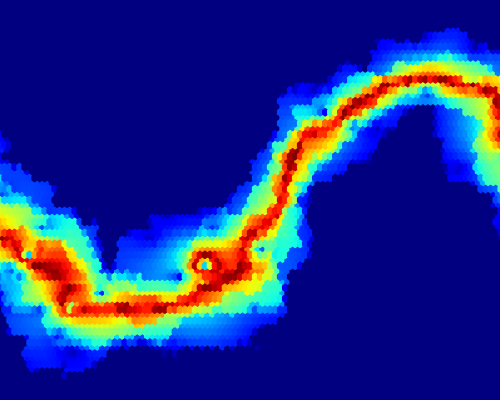}
    \caption{}
    \label{subfig:sin}
    \end{subfigure}
    \begin{subfigure}{.25\linewidth}
    \includegraphics[width=\linewidth]{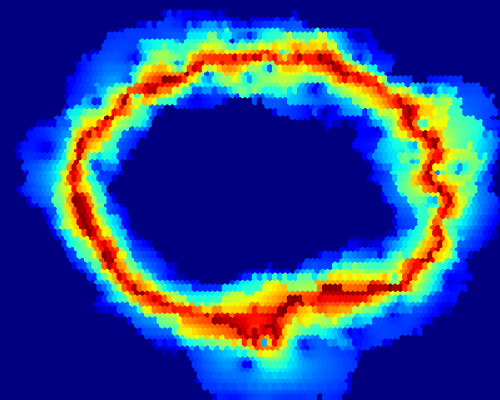}
    \caption{}
    \label{subfig:circle}
    \end{subfigure}
    \caption{Visualization of uncertainty areas in two-dimensional datasets.}\label{fig:datasets}
\end{figure}

Figure~\ref{fig:datasets} represents three two-dimensional datasets, the classes are perfectly separated.
Given the model and one of the uncertainty criteria\footnote{From now on, the model used is $K$-NN ($K$-Nearest Neighbors), with a probabilistic output and on the distance-weighted version available with scikit-learn~\citep{sklearn2011}, every other parameters are scikit-learn default parameters. The uncertainty used is the least confidence measure given  in  equation~\eqref{eq:least_confidence}.}, we can compute the uncertainty of any point in space. For each dataset, the areas of uncertainty of the model are represented, with more red for more uncertainty. It is remarkable that these uncertainty areas can be compared to the decision boundaries of the model.
Often, the closer the observation is to the decision boundary, the less confident the model is about its prediction.

Uncertainty sampling consists of choosing the observation for which the model is the least certain of its prediction. This is one of the basis of active learning, however, other methods allow to extract more information about this uncertainty which leads to the decomposition into epistemic and aleatoric uncertainties.

\section{On the interest and limits of epistemic and aleatoric uncertainties for active learning}\label{section:epistemic}

In this section, we introduce additional elements to decompose the uncertainty of the model so it can focus, in active learning, on the observations that will make it rapidly gain in performance.

\begin{figure}
    \centering
    \begin{subfigure}[t]{.44\linewidth}
    \centering
    \includegraphics[width=0.20\linewidth]{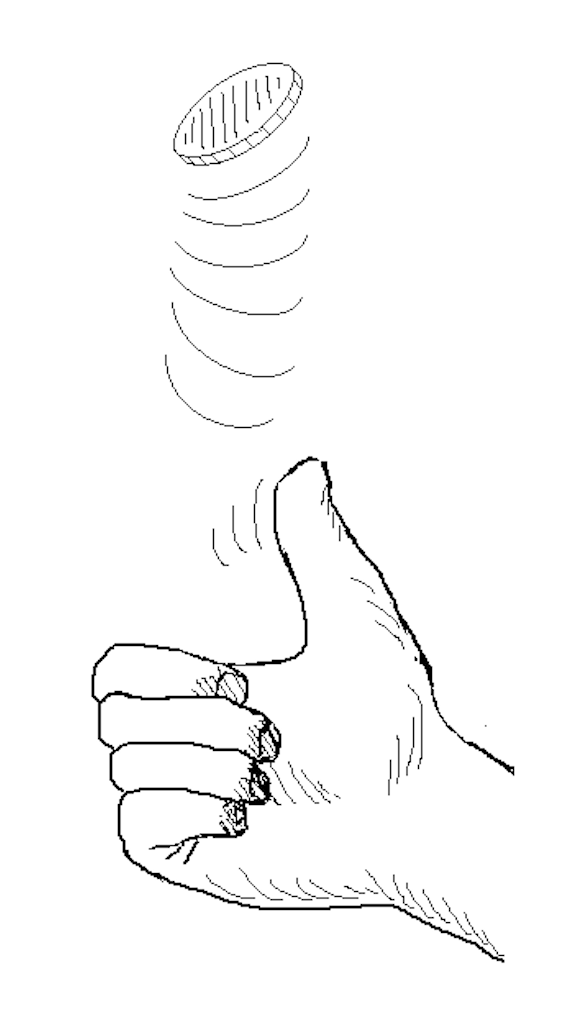}
    \caption{Aleatoric uncertainty}
    \label{subfig:coin}
    \end{subfigure}
    \begin{subfigure}[t]{.44\linewidth}
    \centering
    \includegraphics[width=0.40\linewidth]{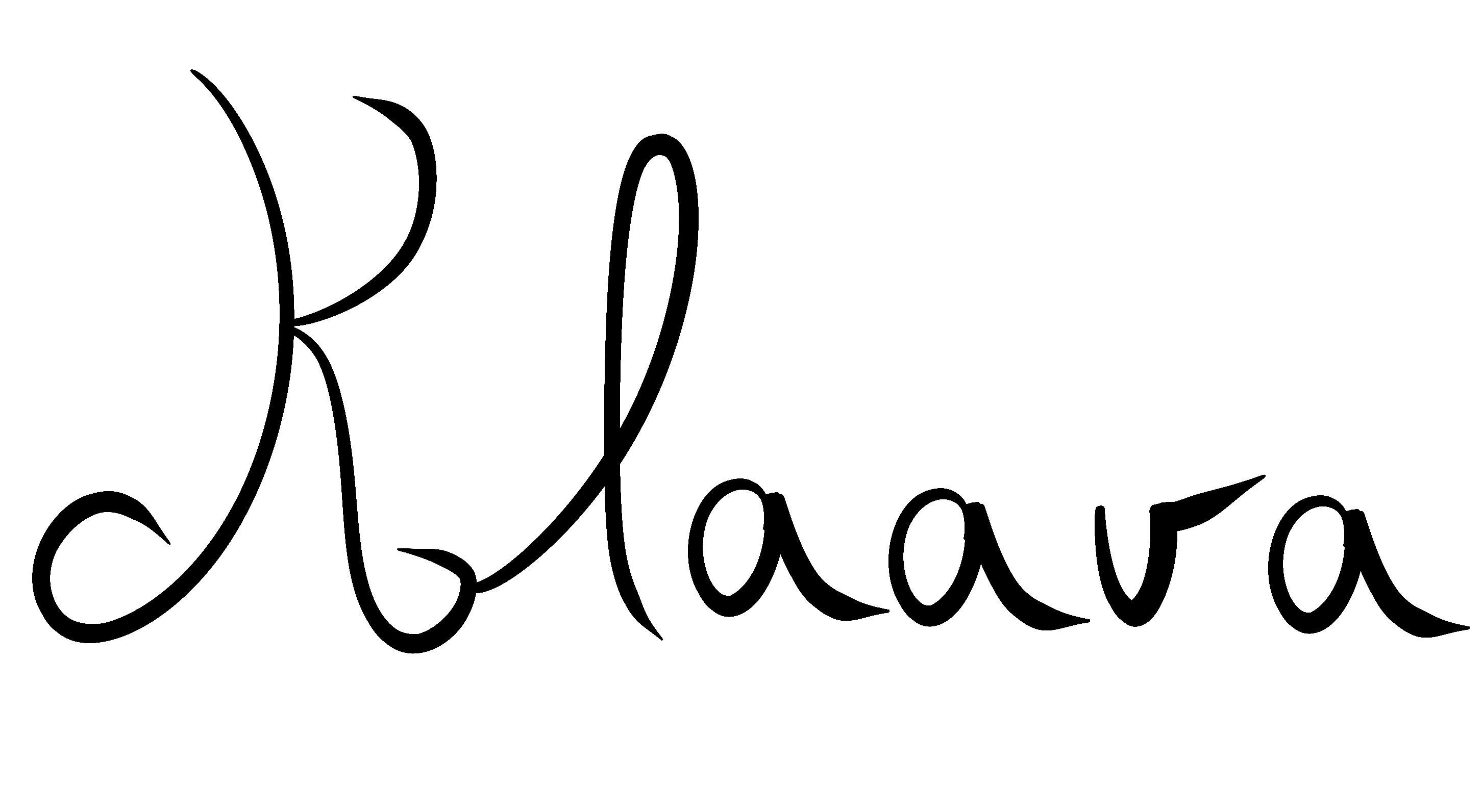}
    \caption{Epistemic uncertainty}
    \label{subfig:word}
    \end{subfigure}
    \caption{Illustration of reducible and irreducible uncertainties in a coin toss experiment (and a Finnish word representation).}\label{fig:coin}
\end{figure}

The uncertainty $\mathcal{U}(x)$ can be divided into two types, as outlined by~\citep{Hora1996}: one is reducible, and the other is irreducible. 
The example provided in Figure~\ref{fig:coin} illustrates these distinctions. In Fig.~\ref{subfig:coin}, the outcome of a coin toss is uncertain, and it is impossible to gain further knowledge to predict whether the coin will land heads or tails. This lack of knowledge is referred to as aleatoric uncertainty. On the other hand, in Fig.~\ref{subfig:word}, a word in Finnish\footnote{In the example, the word ``tails'' is written in Finnish, the word ``heads'' is called \emph{Kruuna}.} representing either heads or tails is shown. This uncertainty can be resolved by learning the language, making it epistemic uncertainty.

Being able to model these two uncertainties can help to delimit where it is more interesting to provide knowledge and where it is useless.
The total uncertainty $\mathcal{U}(x)$ is often represented as the sum of the epistemic uncertainty $\mathcal{U}_e(x)$ and the aleatoric uncertainty $\mathcal{U}_a(x)$: $\mathcal{U}(x) = \mathcal{U}_e(x) + \mathcal{U}_a(x)$.

In a two-class problem where $\Omega = \{0, 1\}$, it is suggested by~\citep{senge2014} to model this uncertainty, here under the formalism of~\citep{Nguyen2022}, by computing the plausibility $\pi$ of belonging to each of the two classes with the following formula, based on a probabilistic model $\theta$:
\begin{equation}\label{equation:ep_plaus}
    \begin{split}
        \pi(1|x) = \underset{\theta \in \Theta}{\text{sup}}\;\text{min} [\pi_\Theta(\theta), p_\theta(1|x) - p_\theta(0|x)],\\
        \pi(0|x) = \underset{\theta \in \Theta}{\text{sup}}\;\text{min} [\pi_\Theta(\theta), p_\theta(0|x) - p_\theta(1|x)],
    \end{split}
\end{equation}
with $\pi_\Theta(\theta)$ depending on the likelihood $L(\theta)$ and the maximum likelihood $L(\hat{\theta})$: 
\begin{equation}
    \pi_\Theta(\theta) = \frac{L(\theta)}{L(\hat{\theta})}.
\end{equation}

The epistemic uncertainty is then high when the two classes are very plausible\footnote{The notion of plausibility within the theory of belief functions used in the proposed methods differs from the one presented here and will be discussed in greater detail in section~\ref{section:method}.} while the aleatoric uncertainty is high when the two classes are implausible:
\begin{equation}\label{eq:ep}
    \begin{split}
        &\mathcal{U}_e(x) = \text{min}[\pi(1|x),\pi(0|x)],\\
        &\mathcal{U}_a(x) = 1 - \text{max}[\pi(1|x),\pi(0|x)].
    \end{split}
\end{equation}

This calculation depends not only on the prediction of the model but also on the observations. To summarize, the fewer observations there are in a region,
or the fewer decision elements there are to strongly predict a class, the higher the plausibility of the two classes, and the more reducible (and thus epistemic) the uncertainty is by adding knowledge. 

\begin{figure}
    \centering
    \begin{subfigure}{.27\linewidth}
    \includegraphics[width=\linewidth]{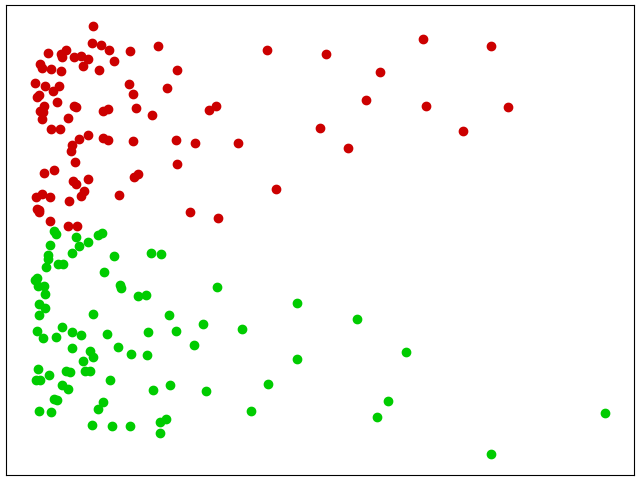}
    \caption{Sample}
    \label{subfig:log}
    \end{subfigure}
    \begin{subfigure}{.25\linewidth}
    \includegraphics[width=\linewidth]{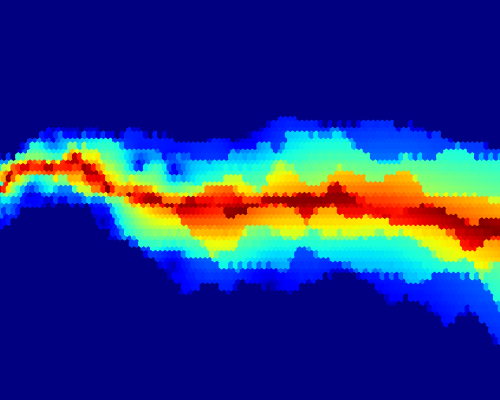}
    \caption{Uncertainty}
    \label{subfig:log_unc}
    \end{subfigure}\\
    \begin{subfigure}{.25\linewidth}
    \includegraphics[width=\linewidth]{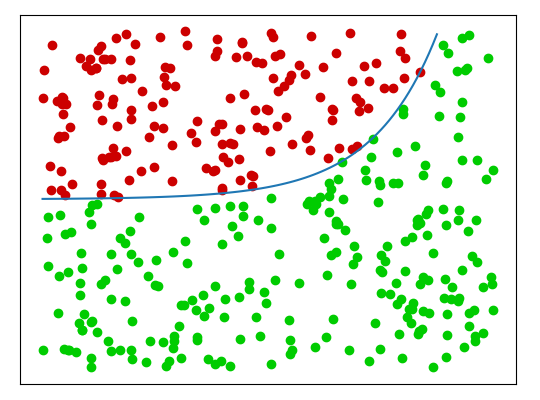}
    \label{subfig:log_top}
    \end{subfigure}
    \begin{subfigure}{.25\linewidth}
    \includegraphics[width=\linewidth]{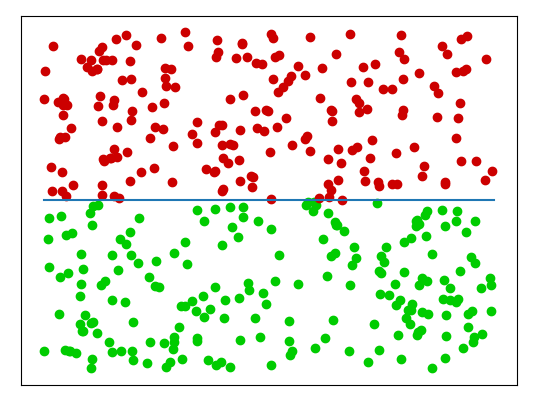}
    \label{subfig:log_middle}
    \end{subfigure}
    \begin{subfigure}{.25\linewidth}
    \includegraphics[width=\linewidth]{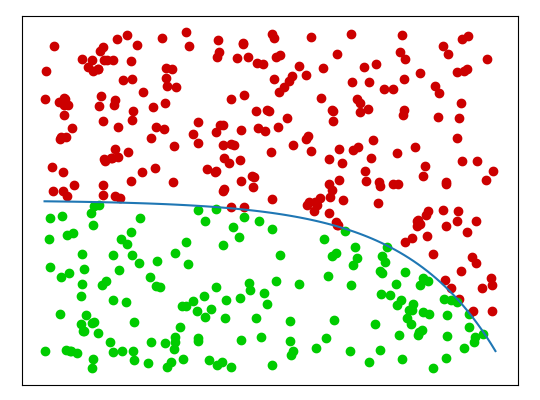}
    \label{subfig:log_bottom}
    \end{subfigure}
    \caption{Visualization of model uncertainty and sample evolution in two-class datasets.}\label{fig:log}
\end{figure}
\begin{figure}
    \centering
    \begin{subfigure}{.25\linewidth}
    \includegraphics[width=\linewidth]{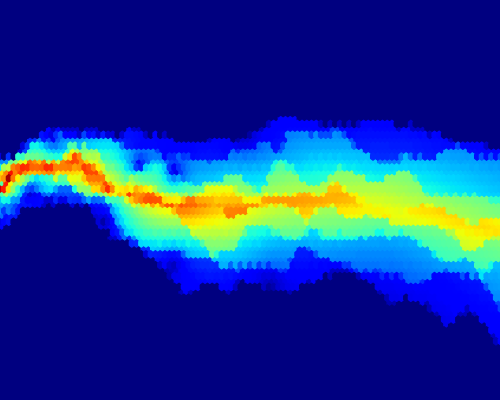}
    \caption{Aleatoric: $\mathcal{U}_a$}
    \label{subfig:log_al}
    \end{subfigure}
    \begin{subfigure}{.25\linewidth}
    \includegraphics[width=\linewidth]{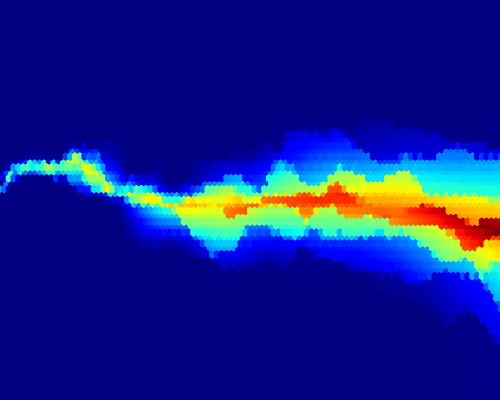}
    \caption{Epistemic: $\mathcal{U}_e$}
    \label{subfig:log_ep}
    \end{subfigure}
    \begin{subfigure}{.25\linewidth}
    \includegraphics[width=\linewidth]{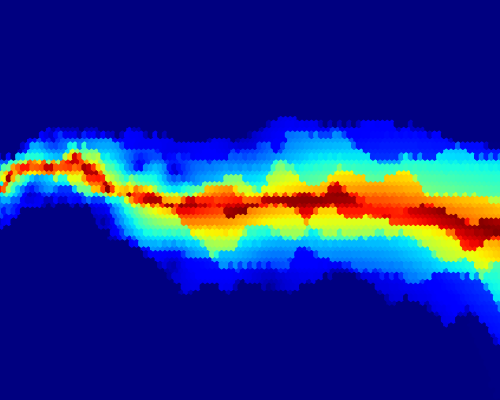}
    \caption{Total: $\mathcal{U}$}
    \label{subfig:log_tot}
    \end{subfigure}
    \caption{Representation of aleatoric and epistemic uncertainties in model predictions according to Fig.~\ref{subfig:log}.}\label{fig:ev_uncertainties}
\end{figure}

An example is shown in Figure~\ref{fig:log}, a two-class dataset is shown in Fig.~\ref{subfig:log} and the areas of model uncertainty are shown in Fig.~\ref{subfig:log_unc} according to the uncertainty sampling presented in the previous section. 

An horizontal line can be distinguished where the model uncertainty is the highest. However, the sample represented in Fig.~\ref{subfig:log}, shows that part of the uncertainty can be removed more easily by adding observations.
In the same figure, three different datasets show how the sample can evolve by adding observations. Whatever the final distribution is, the uncertainty on the left is not very reducible, while the uncertainty on the right can be modified by adding knowledge. 

These two uncertainties can be calculated using equation~\eqref{eq:ep}, and are shown in Figure~\ref{fig:ev_uncertainties}.
The aleatoric uncertainty, and therefore irreducible, is represented in Fig.~\ref{subfig:log_al} and the epistemic uncertainty, reducible, is represented in Fig.~\ref{subfig:log_ep}. The total uncertainty is then the sum of the two, Fig.~\ref{subfig:log_tot}.
Here the goal is to only use the epistemic uncertainty, to know the areas where the model can learn new knowledge and where it will have more impact.

Using epistemic uncertainty as a sampling strategy is not reductive since it theoretically provides similar areas of uncertainty to those used previously when epistemic and aleatoric uncertainties are indistinguishable. But this statement is based on the sum decomposition of total uncertainty, which has recently been questioned.

This information is valuable for identifying areas of reducible uncertainty. However, it is not compatible with richer labels containing uncertainty. Computing this epistemic uncertainty also relies on observations in addition to the model. Essentially, the model defines its zones of uncertainty and seeks locations with the fewest observations to define the reducible uncertainty. Moreover, the exploration-exploitation problem is not fully addressed. This leads to the next section where two uncertainty sampling strategies for rich labels are proposed, extending to multiple classes.

\section{Richer labels and multiple classes}\label{section:method}

In this section, we propose two uncertainty sampling strategies with a simplified calculation phase, capable of handling richer labels. These strategies are no longer directly dependent on observations but only on the model prediction\footnote{The uncertainty no longer depends on observations, but the model does.}. We also propose a natural extension for a number of classes higher than two. The first method uses discord and non-specificity to map uncertainty in order to address the exploration-exploitation problem. The second method extends the epistemic and aleatoric uncertainties to rich labels, also simplifying the computation phase.

\begin{figure}[h]
    \centering
    \includegraphics[width=0.25\linewidth]{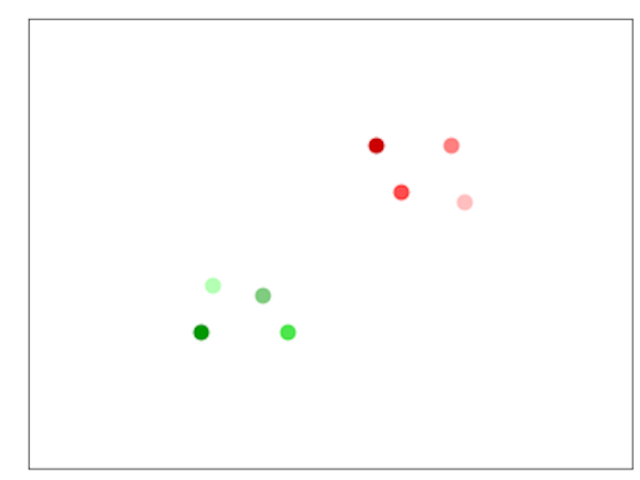}
    \caption{Rich label representation: observations on two dimensions with varying ignorance.}\label{fig:labels}
\end{figure}

From there, a label can be uncertain and imprecise, which means that additional information on ignorance is represented. 
Figure~\ref{fig:labels} illustrates how labels are represented in this document: the darker the dot, the less ignorance the label contains (\emph{e.g., I'm sure this is a dog}); the lighter the dot, the more ignorance it contains (\emph{e.g., I have no idea between dog and cat}). It is important to note that labels are no longer ``hard'', but modeled by a belief function, which allows such a representation.

\subsection{Discord and non-specificity: Klir uncertainty}

In the framework of belief functions, discord and non-specificity are tools that allow to model uncertainty, we propose to use~\citep{Klir1998}'s representation for uncertainty sampling, with potential connections to epistemic and aleatoric uncertainty.

\paragraph{Discord}It is here applied to the output of a model capable of making an uncertain and imprecise prediction\footnote{From now, the Evidential $K$-nearest Neighbors model of~\citep{denoeux1995} is considered.}. Discord represents the amount of conflicting information in the model's prediction. It is, for most models, at its maximum closest to the decision boundary and is calculated using the following formula:
\begin{equation}\label{eq:discord}
    D(m) = -\sum_{A\subseteq\Omega}m(A)\text{ log}_2(BetP(A)),
\end{equation}
with $m$ a mass function, or the output of the model (see section~\ref{section:dempster}).

\begin{figure}[h]
    \centering
    \begin{subfigure}{.25\linewidth}
    \centering
    \includegraphics[width=\linewidth]{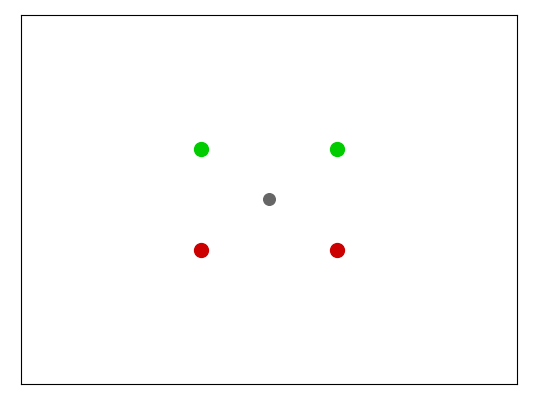}
    \caption{High discord}
    \label{subfig:discord1}
    \end{subfigure}\hspace{0.02\linewidth}
    \begin{subfigure}{.25\linewidth}
    \centering
    \includegraphics[width=\linewidth]{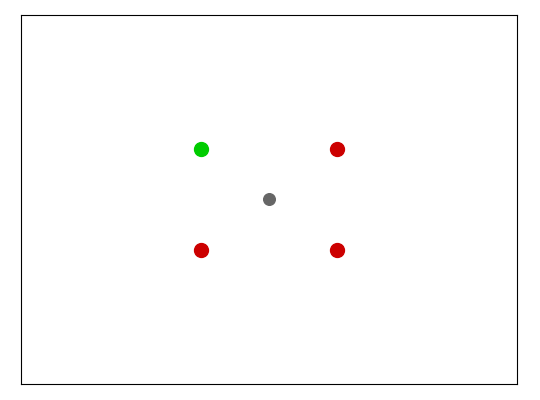}
    \caption{Low discord}
    \label{subfig:discord2}
    \end{subfigure}\hspace{0.02\linewidth}
    \begin{subfigure}{.25\linewidth}
    \centering
    \includegraphics[width=\linewidth]{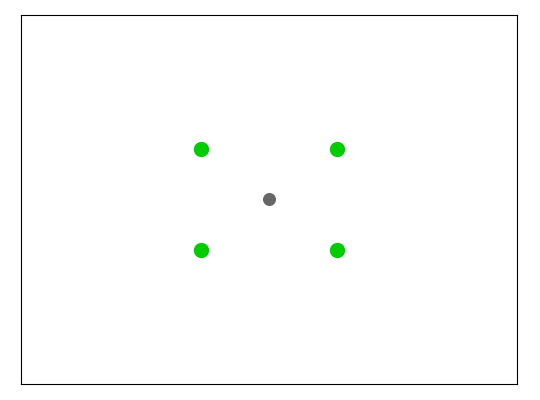}
    \caption{No discord}
    \label{subfig:discord3}
    \end{subfigure}\\
    \begin{subfigure}{.25\linewidth}
    \centering
    \includegraphics[width=\linewidth]{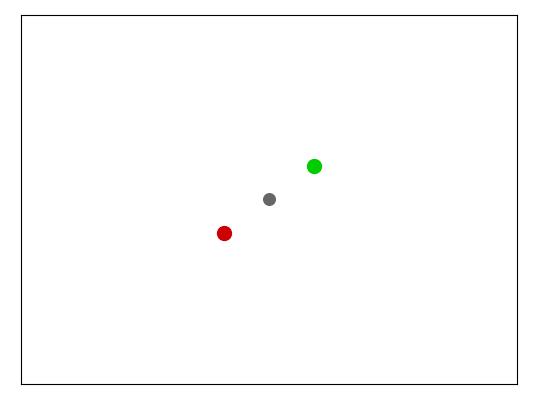}
    \caption{Low non-spec.}
    \label{subfig:nonspe1}
    \end{subfigure}\hspace{0.02\linewidth}
    \begin{subfigure}{.25\linewidth}
    \centering
    \includegraphics[width=\linewidth]{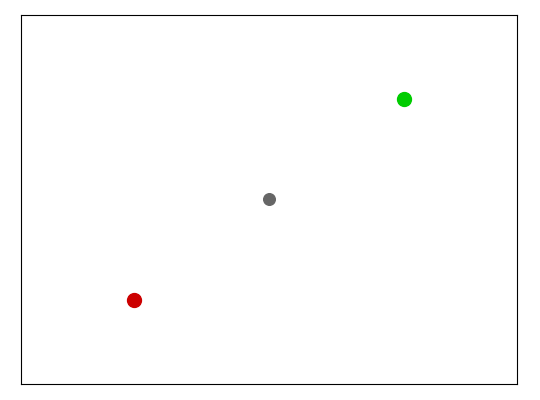}
    \caption{High non-spec.}
    \label{subfig:nonspe2}
    \end{subfigure}\hspace{0.02\linewidth}
    \begin{subfigure}{.25\linewidth}
    \centering
    \includegraphics[width=\linewidth]{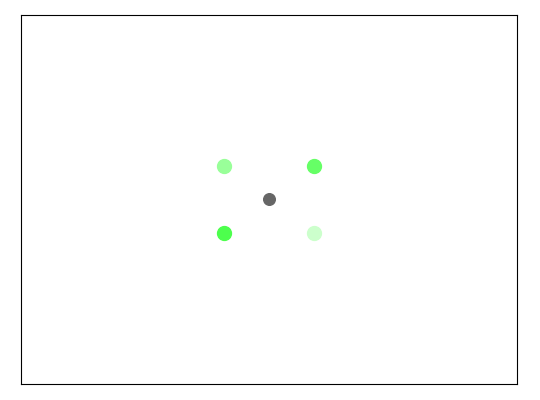}
    \caption{High non-spec.}
    \label{subfig:nonspe3}
    \end{subfigure}
    \caption{Quantifying discord and non-specificity in model uncertainty at the central point.}\label{fig:nonspe}
\end{figure}

Figure~\ref{fig:nonspe} illustrates three different cases where discord varies: from high discordance, where labels around the central point (the observation to label) highly disagree in Fig.~\ref{subfig:discord1}, to low discordance, where each label is in agreement in Fig.~\ref{subfig:discord3}.

\paragraph{Non-Specificity}Non-Specificity quantifies the degree of imprecision of the model~\citep{dubois1987}. This information may be inferred because the model lacks data or because the oracle labeling the instances is itself ignorant. The higher it is, the more imprecise the model's response, it is calculated with:
\begin{equation}\label{eq:nonspe}
    N(m) = \sum_{A\subseteq\Omega}m(A)\text{ log}_2(|A|).
\end{equation}

The same Figure~\ref{fig:nonspe} also represents three different cases of non-specificity, in Fig.~\ref{subfig:nonspe1} the non-specificity is low as there are relevant sources of information next to the observation to be labeled, in Fig.~\ref{subfig:nonspe2} the non-specificity increases the further away the elements are from the observation and in Fig.~\ref{subfig:nonspe3} the non-specificity is also high because the nearby sources of information are themselves ignorant.

\paragraph{Klir uncertainty}This uncertainty is derived from discord and non-specificity, it is used here for uncertainty sampling by adding the two previous formulas:
\begin{equation}
    \mathcal{U}_m(x) = N(x) + D(x),
\end{equation}
with $N(x)$ and $D(x)$ respectively the non-specificity and discord of the model in $x$. \citep{Klir1998} propose to use the same weight for discord and non-specificity, but~\citep{Denoeux2000} introduce a parameter $\lambda\in[0,1]$ that allows to bring more weight to non-specificity (we propose to use it for more exploration) or to discord (for more exploitation):
\begin{equation}\label{eq:ev_unc}
    \mathcal{U}_m(x) = \lambda N(x) + (1-\lambda) D(x).
\end{equation}
Note that this uncertainty is naturally extended to $|\Omega|\geq 2$ classes.

\begin{figure}
    \centering
    \begin{subfigure}{.273\linewidth}
    \includegraphics[width=\linewidth]{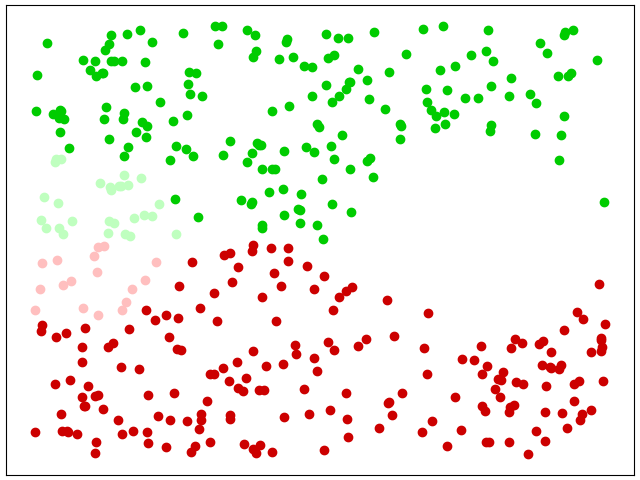}
    \caption{Dataset}
    \label{subfig:imp}
    \end{subfigure}
    \begin{subfigure}{.25\linewidth}
    \includegraphics[width=\linewidth]{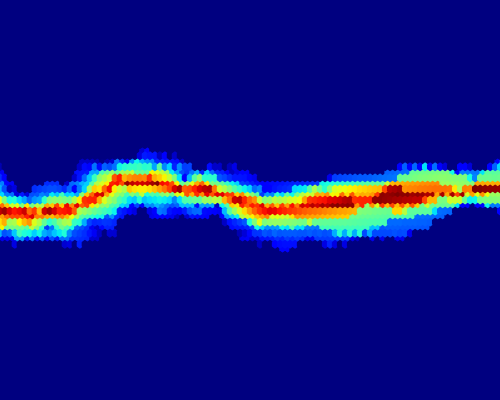}
    \caption{Uncertainty}
    \label{subfig:imp_unc}
    \end{subfigure}
    \begin{subfigure}{.25\linewidth}
    \includegraphics[width=\linewidth]{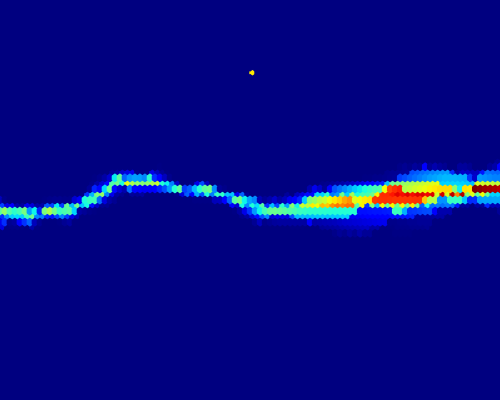}
    \caption{Epistemic}
    \label{subfig:imp_ep}
    \end{subfigure}\\
    \begin{subfigure}{.25\linewidth}
    \includegraphics[width=\linewidth]{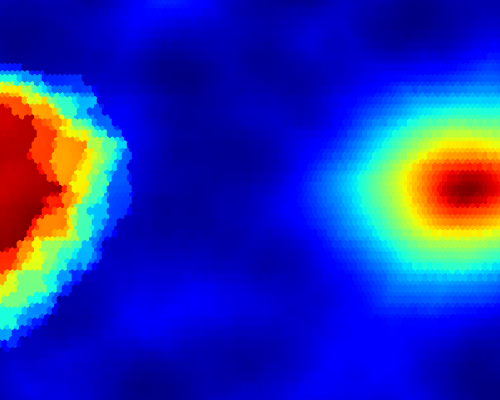}
    \caption{Non-specificity}
    \label{subfig:imp_nonspe}
    \end{subfigure}
    \begin{subfigure}{.25\linewidth}
    \includegraphics[width=\linewidth]{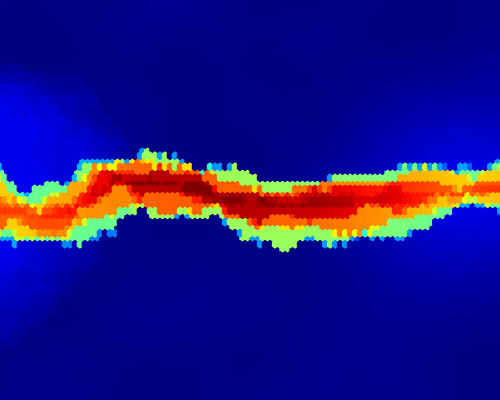}
    \caption{Discord}
    \label{subfig:imp_discord}
    \end{subfigure}
    \begin{subfigure}{.25\linewidth}
    \includegraphics[width=\linewidth]{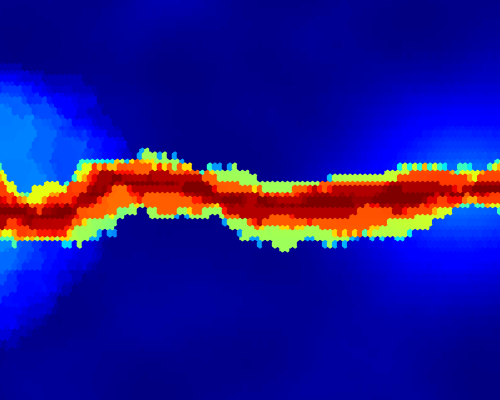}
    \caption{Klir}
    \label{subfig:imp_ev_unc}
    \end{subfigure}\\
    \begin{subfigure}{.25\linewidth}
    \includegraphics[width=\linewidth]{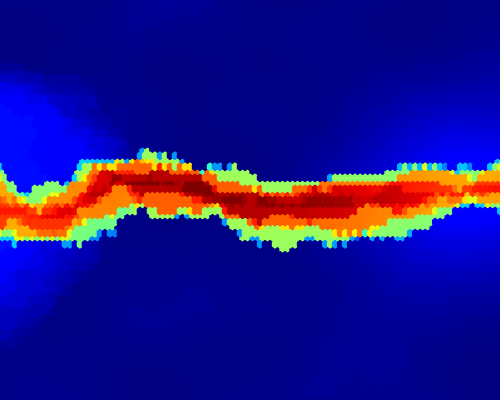}
    \caption{+ Exploitation}
    \label{subfig:more_discord}
    \end{subfigure}
    \begin{subfigure}{.25\linewidth}
    \includegraphics[width=\linewidth]{images/imp_ev_unc.png}
    \caption{Compromise}
    \label{subfig:same}
    \end{subfigure}
    \begin{subfigure}{.25\linewidth}
    \includegraphics[width=\linewidth]{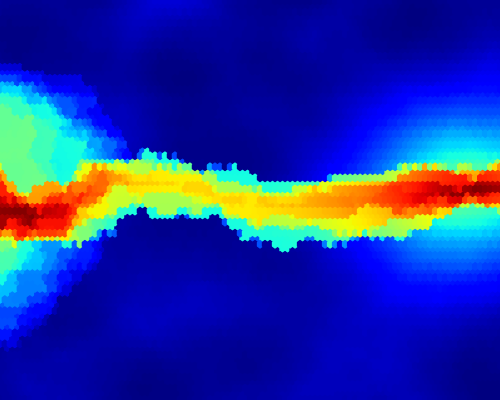}
    \caption{+ Exploration}
    \label{subfig:more_nonspe}
    \end{subfigure}
    \caption{An imperfectly labeled dataset: exploring uncertainty areas through sampling strategies, epistemic uncertainty, and proposed non-specificity, discord, and total Klir uncertainty, alongside the potential for exploration-exploitation compromise.}\label{fig:imp}
\end{figure}

This formula has the advantage of identifying the total uncertainty as well as the reducible one, but also of taking into account the uncertainty already present in the labels and of being adjustable for more exploration or exploitation.
Figure~\ref{fig:imp} depicts a dataset with two areas of uncertainty: on the right, an area with a lack of data, and on the left, an area where labels are more ignorant.
The uncertainty sampling, using Shannon's entropy~\eqref{eq:entropy} or the least confidence measure~\eqref{eq:least_confidence} is not able to see either of these two areas, Fig.~\ref{subfig:imp_unc}.
The epistemic uncertainty~\eqref{eq:ep} is able to distinguish the uncertainty related to the arrangement of the observations in space (\emph{i.e.} the uncertainty on the right) but not the uncertainty related to the ignorance of the sources, Fig.~\ref{subfig:imp_ep}.

The proposal of using Klir uncertainty for sampling (discord and non-specificity) allows to represent each of these uncertainties. The second line shows the areas of non-specificity Fig.~\ref{subfig:imp_nonspe}, of discord Fig.~\ref{subfig:imp_discord} and Klir uncertainty Fig.~\ref{subfig:imp_ev_unc}.

Klir uncertainty can then be used for uncertainty sampling in active learning, it is also possible to vary the result for more exploration or more exploitation by modifying~$\lambda$. The last line shows the areas of uncertainty for different values of~$\lambda$, more discord on the left Fig.~\ref{subfig:more_discord} to more non-specificity on the right Fig.~\ref{subfig:more_nonspe}.

We have proposed here to use Klir's uncertainty in sampling, which allows to represent some uncertainties areas in active learning related to rich labels. The method is no longer dependent on the observations, but only on the prediction of the model and the exploration-exploitation problem is addressed thanks to the $\lambda$ parameter. Even though discord may recall aleatoric uncertainty (non-reducible) and non-specificity may recall epistemic uncertainty (reducible), these notions are not equivalent. Therefore, in the following section we also propose an extension of epistemic (and aleatoric) uncertainty for rich labels and for several classes.

\subsection{Evidential epistemic uncertainty}

We propose here to extend the notion of epistemic uncertainty (\emph{c.f.} section~\ref{section:epistemic}) to rich labels (\emph{c.f.} section~\ref{section:rich_labels}), by removing the dependence on observations, simplifying the computational phase, and allowing the model to detect new areas of uncertainty.

The epistemic uncertainty can be extended to rich labels by using the notion of plausibility within the framework of belief functions (which differs here from the one presented in section~\ref{section:epistemic}). It represents the total evidence that does not support the complementary event for a class $\omega$ or more generally for an element $A\in 2^\Omega$.
The plausibility $Pl$ defines the belief that could be allocated to $A$:
\begin{equation}
  Pl(A) = \sum_{A\cap B\neq\emptyset}m(B).
\end{equation}
The plausibility being the consistent evidence, the belief function $Bel$ defines the total evidence directly supporting $A$:
\begin{equation}
  Bel(A) = \sum_{B\subseteq A, B \neq\emptyset}m(B).
\end{equation}
We have $Pl(A) = 1 - Bel(\bar{A})$. For example, suppose the only reliable evidence is that a picture depicts an animal, and a Dog is an animal. In this scenario, it is entirely plausible that the picture is a Dog (plausibility is 1), yet there is no direct belief that the picture is a Dog (belief is 0).
Analogous to equation~\eqref{eq:ep} and for two classes $\Omega = \{0, 1\}$ the epistemic uncertainty is maximal when both classes are highly plausible. 
The proposed evidential epistemic and aleatoric uncertainties are defined as follows:
\begin{equation}
    \begin{split}
        &\mathcal{U}_e(x) = \text{min}[Pl(1|x), Pl(0|x)],\\
        &\mathcal{U}_a(x) = 1 - \text{max}[Pl(1|x), Pl(0|x)].
    \end{split}
\end{equation}
The equation for the aleatoric uncertainty can be rewritten depending on the belief $Bel$:
\begin{equation}
        \mathcal{U}_a(x) = \text{min}[Bel(1|x), Bel(0|x)].
\end{equation}
The sum of the epistemic and aleatoric uncertainties is then the total evidential uncertainty: $\mathcal{U}(x) = \mathcal{U}_e(x) + \mathcal{U}_a(x)$.
However, when the number of classes exceeds~2 the equation of the epistemic uncertainty cannot be simplified by the minimum plausibility:
\begin{equation}
    \begin{split}
        &\mathcal{U}_e(x) \neq \text{min}([Pl(\omega|x)|\omega \in \Omega]),\\
        &\mathcal{U}_a(x) \neq 1 - \text{max}([Pl(\omega|x)|\omega \in \Omega]).
    \end{split}
\end{equation}

It is preferable to first define the uncertainty related to one of the classes $\omega$, rewritten with the belief $Bel$ to avoid having to manipulate $\bar{\omega}$: 
\begin{equation}
    \begin{split}
        \mathcal{U}_e(\omega|x) &= \text{min}[Pl(\omega|x), Pl(\bar{\omega}|x)]\\
        &= \text{min}[Pl(\omega|x), 1 - Bel(\omega|x)].
    \end{split}
\end{equation}
The evidential extension of the epistemic and aleatoric uncertainties for $|\Omega| \geq 2$ classes is then:
\begin{equation}\label{eq:ev_ep}
    \begin{split}
        &\mathcal{U}_e(x) = \sum_{\omega\in\Omega}\text{min}[Pl(\omega|x), 1 - Bel(\omega|x)],\\
        &\mathcal{U}_a(x) = \sum_{\omega\in\Omega}\text{min}[Bel(\omega|x), 1 - Pl(\omega|x)].
    \end{split}
\end{equation}

\begin{figure}
    \centering
    \begin{subfigure}{.273\linewidth}
    \includegraphics[width=\linewidth]{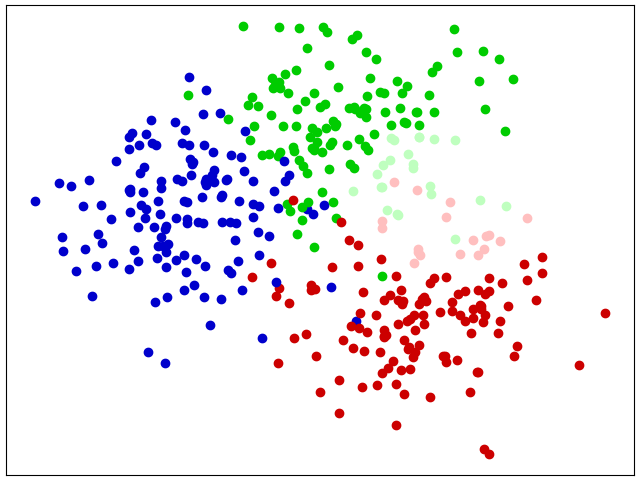}
    \caption{Dataset}
    \label{subfig:triple}
    \end{subfigure}
    \begin{subfigure}{.25\linewidth}
    \includegraphics[width=\linewidth]{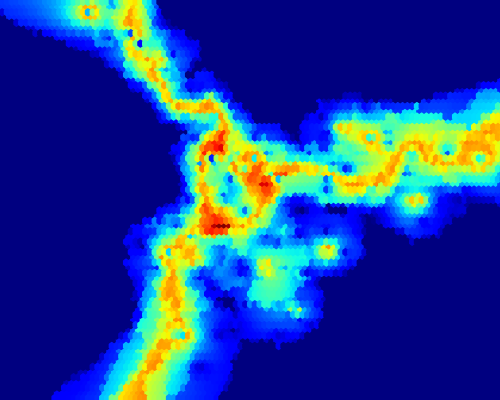}
    \caption{Uncertainty}
    \label{subfig:triple_unc}
    \end{subfigure}\\
    \begin{subfigure}{.25\linewidth}
    \includegraphics[width=\linewidth]{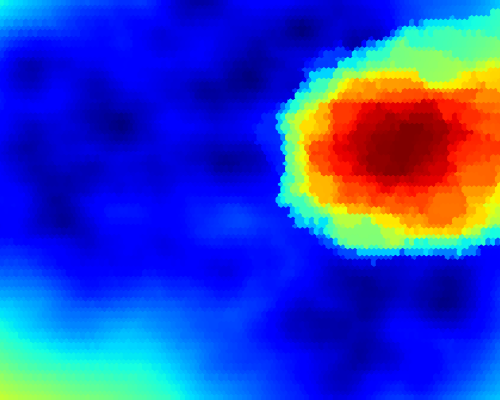}
    \caption{Non-specificity}
    \label{subfig:triple_nonspe}
    \end{subfigure}
    \begin{subfigure}{.25\linewidth}
    \includegraphics[width=\linewidth]{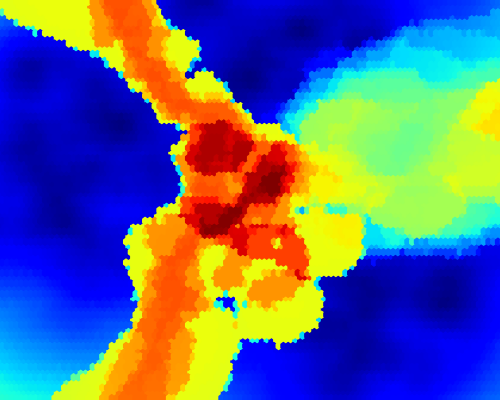}
    \caption{Discord}
    \label{subfig:triple_discord}
    \end{subfigure}
    \begin{subfigure}{.25\linewidth}
    \includegraphics[width=\linewidth]{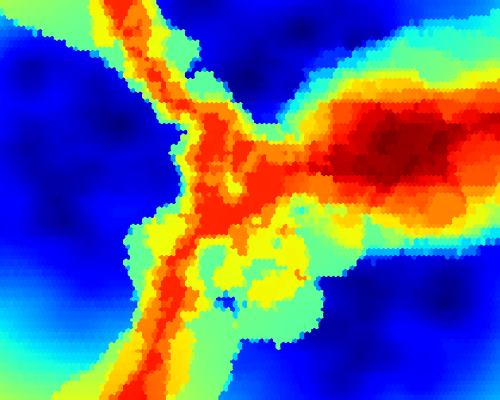}
    \caption{Klir}
    \label{subfig:triple_ev}
    \end{subfigure}\\
    \begin{subfigure}{.25\linewidth}
    \includegraphics[width=\linewidth]{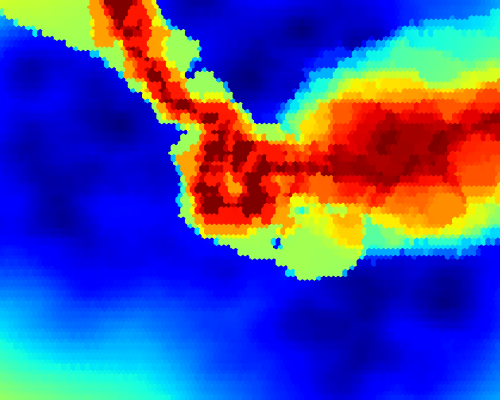}
    \caption{Green class}
    \label{subfig:triple_ev_ep_21}
    \end{subfigure}
    \begin{subfigure}{.25\linewidth}
    \includegraphics[width=\linewidth]{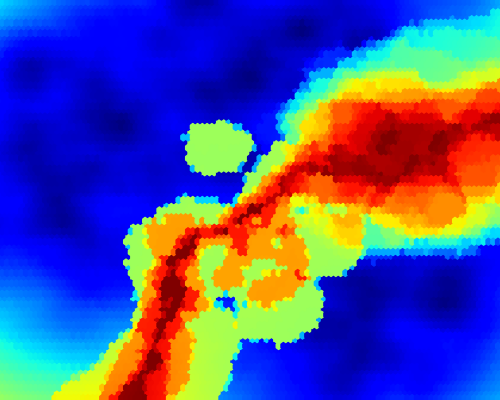}
    \caption{Red class}
    \label{subfig:triple_ev_ep_22}
    \end{subfigure}
    \begin{subfigure}{.25\linewidth}
    \includegraphics[width=\linewidth]{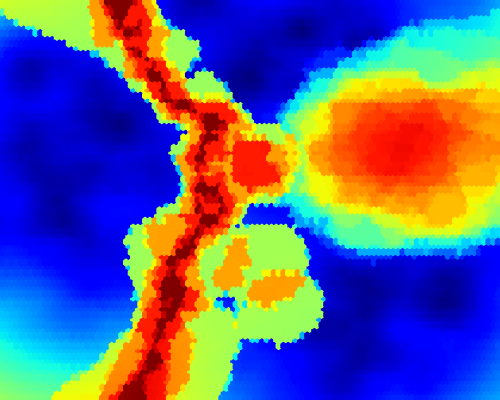}
    \caption{Blue class}
    \label{subfig:triple_ev_ep_23}
    \end{subfigure}\\
    \begin{subfigure}{.25\linewidth}
    \includegraphics[width=\linewidth]{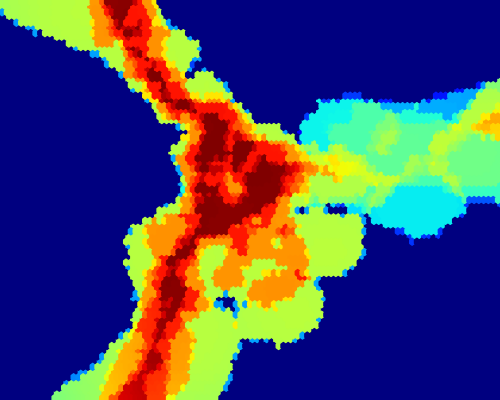}
    \caption{Ev. aleatoric}
    \label{subfig:triple_ev_al}
    \end{subfigure}
    \begin{subfigure}{.25\linewidth}
    \includegraphics[width=\linewidth]{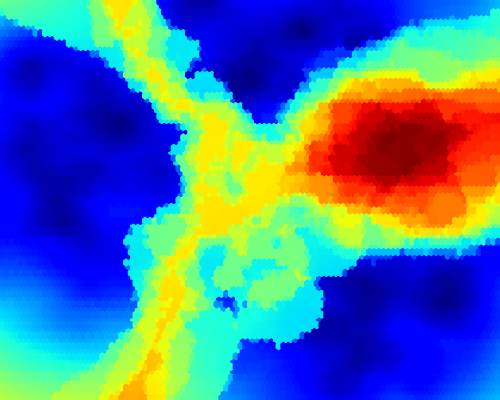}
    \caption{Ev. epistemic}
    \label{subfig:triple_ev_ep}
    \end{subfigure}
    \begin{subfigure}{.25\linewidth}
    \includegraphics[width=\linewidth]{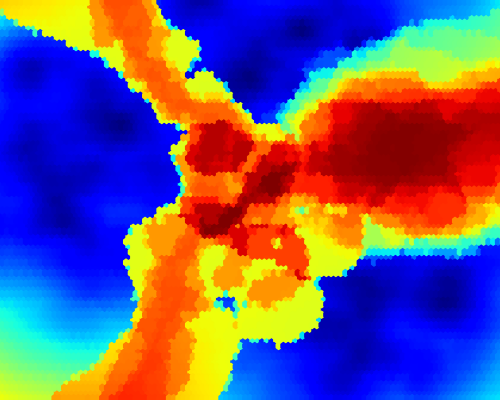}
    \caption{Ev. total}
    \label{subfig:triple_ev_tot}
    \end{subfigure}
    \caption{A three-class dataset: representing label imprecision and uncertainty zones.}\label{fig:triple}
\end{figure}

The example in Figure~\ref{fig:triple} shows a dataset of three classes with a zone of imprecision for some labels (between the green and red classes).
Probabilistic~\eqref{eq:entropy}-\eqref{eq:least_confidence} and epistemic~\eqref{eq:ep} uncertainties cannot model the imprecision present in the labels, this less complete uncertainty zone is represented in Fig.~\ref{subfig:triple_unc}.

The previous uncertainty resulting from the sum of the discord and the non-specificity is also presented. It manages both exploration, Fig.~\ref{subfig:triple_nonspe}, and exploitation, Fig.~\ref{subfig:triple_discord}, to give a better representation of the uncertainty, Fig.~\ref{subfig:triple_ev}.

The extension of the epistemic uncertainty, also introduced in this paper, is presented in the same figure. First, the evidential epistemic areas of uncertainties for each of the three classes are presented in Fig.~\ref{subfig:triple_ev_ep_21}, Fig.~\ref{subfig:triple_ev_ep_22} and Fig.~\ref{subfig:triple_ev_ep_23}.
Then, the resulting evidential epistemic uncertainty of the model is deducted from equation~\eqref{eq:ev_ep} in Fig.~\ref{subfig:triple_ev_ep} along with the evidential aleatoric and total uncertainties.

\section{Experiments}\label{section:experiments}

In this section, we conduct two types of experiments. The first is more theoretical, applying the two proposed methods to a dataset with real rich labels. The second is more traditional in active learning, comparing one of the methods with uncertainty sampling on several real-world datasets. The exploration-exploitation dilemma is addressed in this second part.

This aim of the first exploratory and non-comparative experiment is to demonstrates how information is mapped by the model. The second experiment offers a more traditional metric-based approach in active learning, allowing for a tangible comparison of methods using classical metrics used in this domain~\citep{kottke2017challenges}.

\subsection{Sampling on real world dataset}\label{section:real}

\begin{figure}
    \centering
    \begin{subfigure}{.3\linewidth}
    \includegraphics[width=\linewidth]{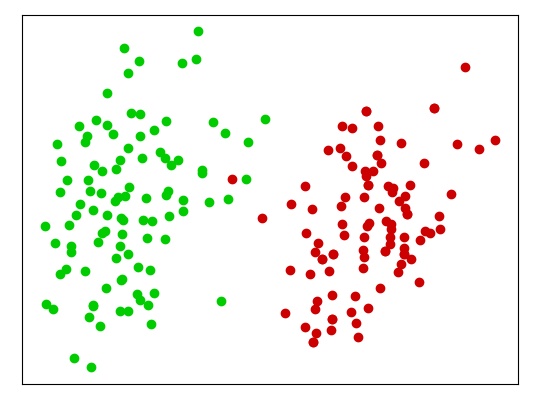}
    \caption{True labels}
    \label{subfig:perfect}
    \end{subfigure}
    \begin{subfigure}{.3\linewidth}
    \includegraphics[width=\linewidth]{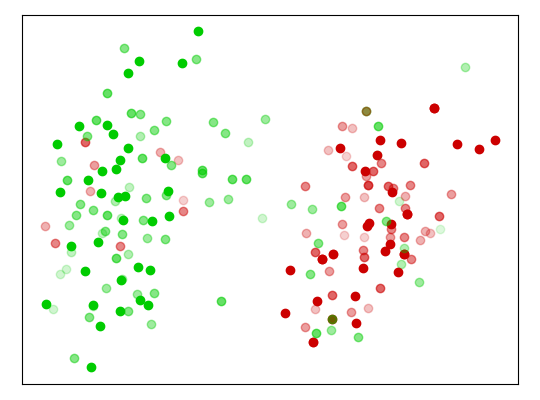}
    \caption{Rich labels}
    \label{subfig:imperfect}
    \end{subfigure}\\
    \begin{subfigure}{.25\linewidth}
    \includegraphics[width=\linewidth]{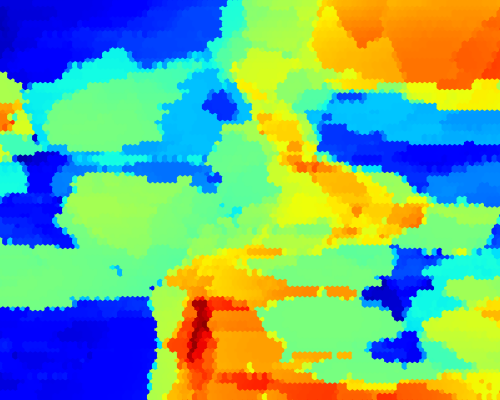}
    \caption{Discord}
    \label{subfig:real_discord}
    \end{subfigure}
    \begin{subfigure}{.25\linewidth}
    \includegraphics[width=\linewidth]{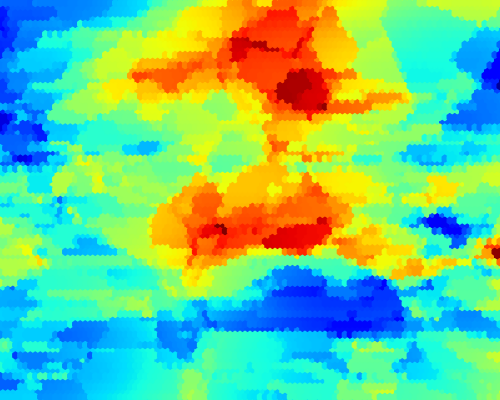}
    \caption{Non-specificity}
    \label{subfig:real_nonspe}
    \end{subfigure}
    \begin{subfigure}{.25\linewidth}
    \includegraphics[width=\linewidth]{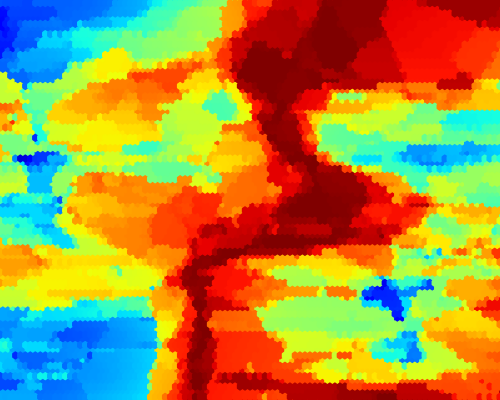}
    \caption{Klir}
    \label{subfig:real_ev_unc}
    \end{subfigure}\\
    \begin{subfigure}{.25\linewidth}
    \includegraphics[width=\linewidth]{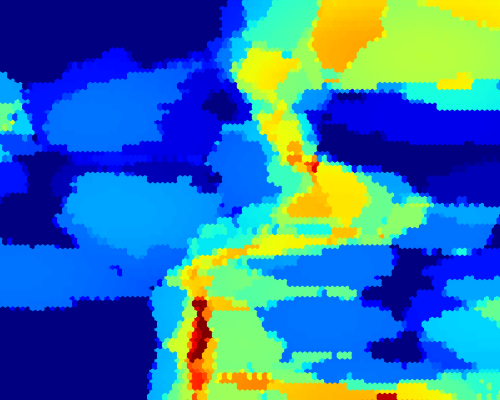}
    \caption{Ev. aleatoric}
    \label{subfig:real_al}
    \end{subfigure}
    \begin{subfigure}{.25\linewidth}
    \includegraphics[width=\linewidth]{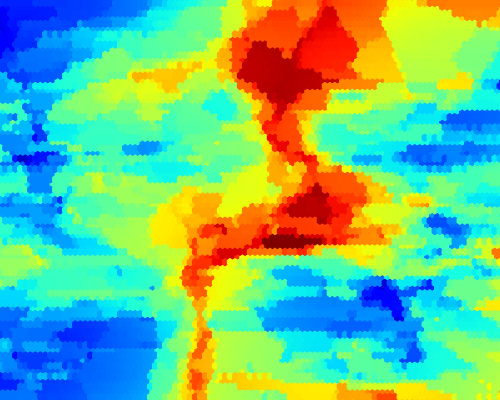}
    \caption{Ev. epistemic}
    \label{subfig:real_ep}
    \end{subfigure}
    \begin{subfigure}{.25\linewidth}
    \includegraphics[width=\linewidth]{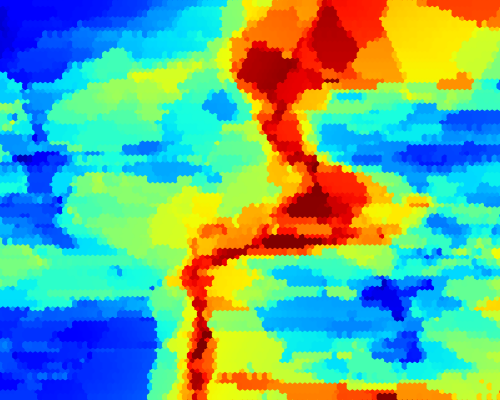}
    \caption{Ev. total}
    \label{subfig:real_tot}
    \end{subfigure}
    \caption{Ignorance mapping in the Credal Dog-2 dataset: Brittany breed (green) and Beagle breed (red).}\label{fig:real}
\end{figure}

In this section we use datasets for which we have access to truly imperfectly labeled data with rich labels. This part is exploratory in nature and does not endorse the superiority of any method. Moreover, conventional methods for computing model uncertainty do not take into account the degrees of imprecision of these rich labels and only have access to hard labels. This paper proposes two methods capable of addressing this gap. They are illustrated on Credal Dog-2, a dataset labeled uncertainly and imprecisely by users during crowdsourcing campaigns~\citep{Hoarau2023-1}. Figure~\ref{fig:real} shows the dataset, on the two first components of a Principal Component Analysis to represent this 42-variable dataset in 2D. This is a two-class dataset represented in Fig.~\ref{subfig:perfect} with true classes and in Fig.~\ref{subfig:imperfect} with uncertain and imprecise rich labels given by contributors. Darker dots indicate higher certainty, and \emph{vice versa}.

The first proposed method, sampling by Klir uncertainty, is demonstrated on the dataset with rich labels, Fig.~\ref{subfig:imperfect}. The non-specificity is presented in Fig.~\ref{subfig:real_nonspe}, and can be interpreted as the zones of imprecision of the model, either because it has not had enough access to information (lack of data) or because the users who labeled these instances are themselves ignorant. Discord is also represented in Fig.~\ref{subfig:real_discord}, these are the areas where the model's prediction is conflicting, meaning it is closest to its decision boundary. The total uncertainty in Fig.~\ref{subfig:real_ev_unc} is the sum of the two, it is this latter information that is used to sample on the model uncertainty.

The second proposed method extends epistemic uncertainty, which is a reducible uncertainty applied to evidential reasoning. The irreducible aleatoric evidential uncertainty in Fig.~\ref{subfig:real_al} is presented along with the reducible epistemic evidential uncertainty in Fig.~\ref{subfig:real_ep}. The total uncertainty in Fig.~\ref{subfig:real_tot} is the sum of the reducible and irreducible uncertainties. For active learning, it is not the total uncertainty, but the epistemic reducible uncertainty that is used. Similarities can be noted between discord and aleatoric uncertainty and between non-specificity and epistemic uncertainty. Additionally, the areas of total uncertainty are also similar.

One advantage of the methods proposed in this paper is their ability to account for the uncertainty already present in the labels  (\emph{i.e.} the uncertainty of the oracles).
During labeling by human\footnote{This representation also applies to labeling performed by a machine.} (as in the Cedal Dog-2 dataset), a user may hesitate between class 1 and 2. In such cases, it is preferable to model their uncertainty rather than forcing them to provide a wrong label for one of the classes, which would introduce noise into the dataset. The advantage of belief functions is that for multiple classes, the user can respond with various degrees of ignorance. For instance, they may indicate that they are uncertain about the true class but are relatively confident that it is not class 4 or 2. Therefore, being able to represent the imperfections of the oracle can also lead to improved results in machine learning. Unfortunately, extremely expressive visual results are hard to obtain (and interpret) since these datasets are rare, difficult to collect, and very noisy. However, the following experiment, in section~\ref{section:active_learning}, demonstrates that the proposed methods perform very significantly on this dataset.
 
\subsection{Application to active learning}\label{section:active_learning}

Sampling by Klir uncertainty was chosen for this series of experiments, and the only parameter $\lambda$ in the proposed method is at the heart of this study. A $\lambda$ that tends towards $0$ implies more exploitation, whereas a $\lambda$ that tends towards $1$ implies more exploration. For these experiments, we set $\lambda$ to 0.2, which means that the model will opt for more exploitation than exploration (this choice is motivated in section~\ref{section:discussion}).
Below, the results are compared with random sampling (the baseline) and the very popular uncertainty sampling~\eqref{eq:least_confidence}. 

The experiments\footnote{Experiments where conducted according to the following code: \url{https://anonymous.4open.science/r/evidential-active-learning-B266}} have been carried out on datasets containing between 2 and 8 classes with a number of observations in different ranges. We used well-known datasets available on the UCI Machine Learning Repository~\citep{Dua2019}, and very often used in active learning, as well as \emph{Dog-2}, the dataset presented in the previous section. Table~\ref{exp:datasets} describes these datasets, with the number of observations (or instances), the number of classes, the number of features, and the entropy for the distribution of classes\footnote{An entropy of $1$ means that the classes are perfectly equidistributed and an entropy of $0$ would indicate the total over-representation of one of the classes.}.

\begin{table}[h]
    \caption{Datasets description, with number of observations, classes, features and class distribution entropy.}\label{exp:datasets}%
    \begin{tabular}{@{}lcccc@{}}
    \toprule
    Dataset & Observations & Classes & Features & Entropy\\
    \midrule
    Bank & 1372 & 2 & 4 & 0.99\\
    Qsar & 1055 & 2 & 41 & 0.92\\
    Blod & 748 & 2 & 4 & 0.79\\
    Breast Cancer & 569 & 2 & 30 & 0.95\\
    Ionosphere & 351 & 2 & 34 & 0.94\\  
    Heart & 303 & 2 & 7 & 1.00\\
    Liver & 345 & 2 & 6 & 0.98\\
    Sonar & 208 & 2 & 60 & 1.00\\
    Parkinson & 195 & 2 & 22 & 0.81\\
    Dog-2 & 200 & 2 & 42 & 1.00\\
    Seeds & 210 & 3 & 7 & 1.00\\ 
    Iris & 150 & 3 & 4 & 1.00\\ 
    Wine & 178 & 3 & 13 & 0.99\\ 
    Glass & 214 & 6 & 9 & 0.83\\ 
    Ecoli & 336 & 8 & 7 & 0.73\\ 
    \botrule
    \end{tabular}
\end{table}
Since the goal of active learning is to reduce the cost of labeling, one experiment involves evaluating the model's performance as observations are progressively labeled. Experiments are arbitrarily stopped once the dataset has been labeled at $60\%$ (it will be clear from the graphs that there is no point in going any further). The model is the same for each method, the Evidential $K$-Nearest Neighbors of~\citep{denoeux1995}, with $K=7$ neighbours (see~\citep{Hoarau2022} for parameter selection).
Each experiment is performed 100 times to obtain an estimation of the actual mean accuracy of the model for each dataset. Several criteria are used to compare the results, including accuracy, the area under the accuracy curve (AUAC) and the rank obtained for each dataset. For evaluation, several statistical tests are conducted, including Student's t-test for AUACs, Friedman's test, and the Wilcoxon-Holm method for critical difference diagrams~\citep{demvsar2006statistical}.

\begin{figure}[h]%
    \centering
    \hspace{-0.4cm}
    \begin{subfigure}{.32\linewidth}
		\includegraphics[width=1.1\linewidth]{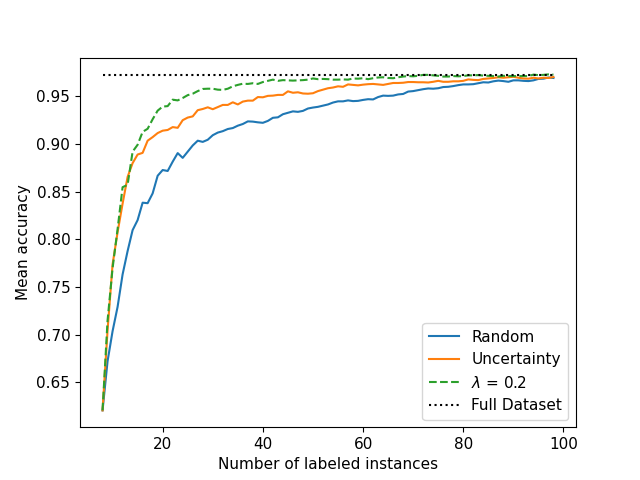}
		\caption{Dog-2}
	\end{subfigure}
    \begin{subfigure}{.32\linewidth}
		\includegraphics[width=1.1\linewidth]{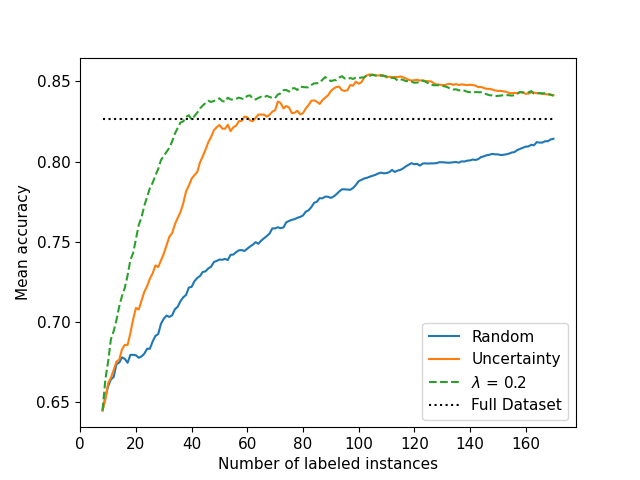}
		\caption{Ionosphere}
	\end{subfigure}
    \begin{subfigure}{.32\linewidth}
		\includegraphics[width=1.1\linewidth]{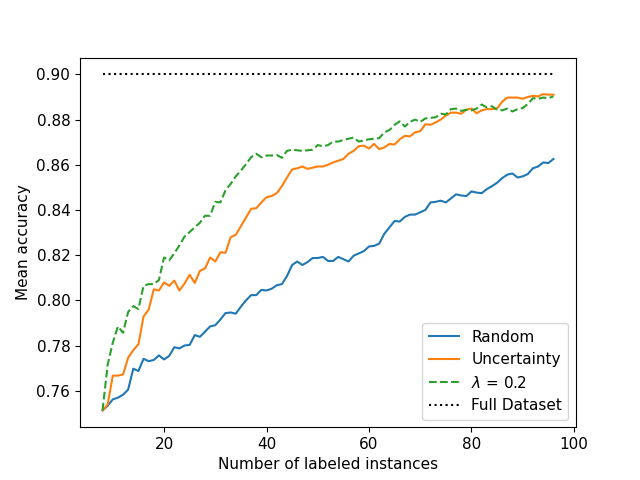}
		\caption{Parkinson}
	\end{subfigure}\\
	\hspace{-0.4cm}
    \begin{subfigure}{.32\linewidth}
		\includegraphics[width=1.1\linewidth]{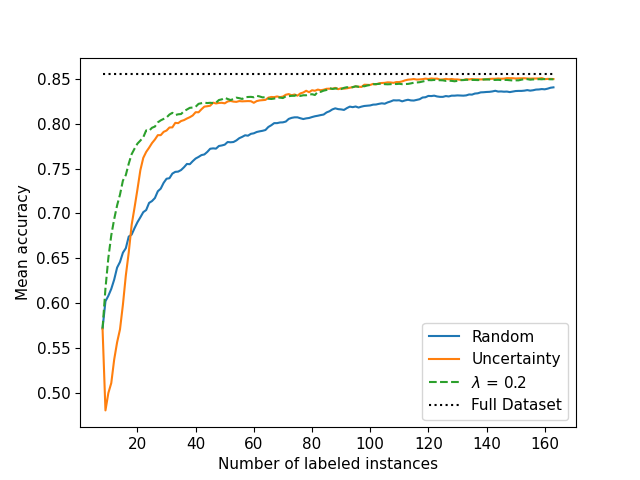}
		\caption{Ecoli}
	\end{subfigure}
    \begin{subfigure}{.32\linewidth}
		\includegraphics[width=1.1\linewidth]{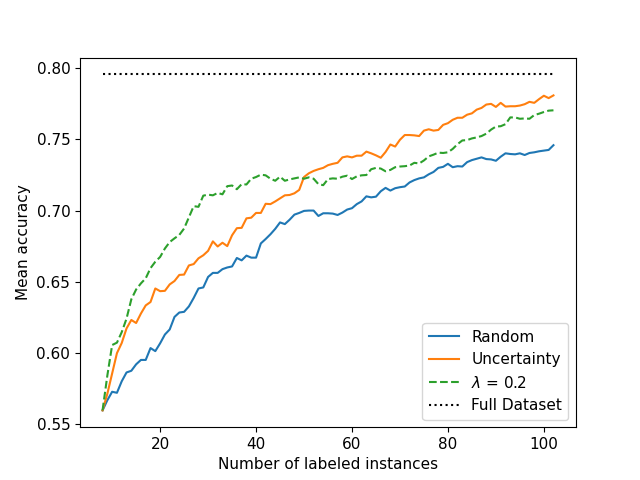}
		\caption{Sonar}
	\end{subfigure}
    \begin{subfigure}{.32\linewidth}
		\includegraphics[width=1.1\linewidth]{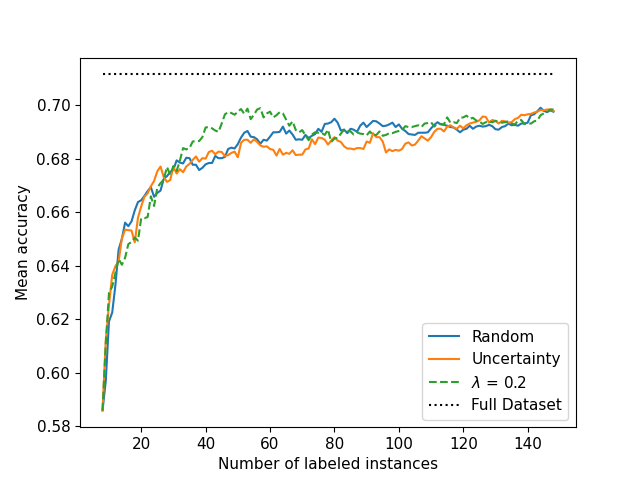}
		\caption{Heart}
	\end{subfigure}
    \caption{Mean accuracy vs. number of labeled instances for Random Sampling, Uncertainty Sampling, and the proposed method with $\lambda = 0.2$ for 6 datasets.}\label{exp:al}
\end{figure}

Figure~\ref{exp:al} shows 6 of the 15 datasets where the proposed method offers the most significant performance, the final performance on the full labeled dataset (\emph{i.e.} if there is no active learning) is represented by the dotted curve and the dashed curve represents the proposed method with $\lambda = 0.2$. In each graph, the superiority of the proposed method over uncertainty sampling is evident, particularly in terms of AUAC. Notably, for the Sonar and Heart datasets, this superiority is only temporary, observed mainly at the beginning of active learning for Sonar and in the middle for Heart. Nevertheless, it will be demonstrated that this domination is not always statistically confirmed, especially for the Parkinson dataset.
Assuming an identical labelling cost for each observation, some insights can be reported below.
\subsubsection*{Some active learning insights on Figure~\ref{exp:al}}
\begin{itemize}
    \item On Dog-2: When reaching $99\%$ of the full dataset performance, uncertainty sampling manages to reduce labeling cost by $62\%$ whereas the proposed method manages to reduce the cost by $82\%$.
    \item On Ionosphere: Using the proposed method, the labeling costs can be reduced by a factor of $9$ with $0\%$ of accuracy loss with respect to the full dataset, whereas with uncertainty sampling, to allow labeling $9$ times cheaper, the model would loose $6\%$ of accuracy (for this dataset, the reduction in labeling cost can improve the performance of the model, a phenomenon which sometimes occurs in active learning, represented by the active learning curve which exceeds the full dataset performance horizontal line).
    \item On Ecoli: It takes $10$ steps for uncertainty sampling to reach the performance of the proposed method after only $3$ steps. 
\end{itemize}

\begin{figure}[h]%
    \centering
    \begin{subfigure}{.32\linewidth}
		\includegraphics[width=\linewidth]{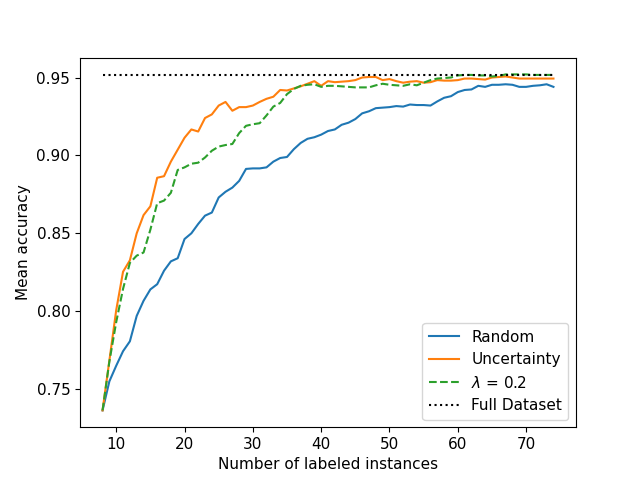}
		\caption{Iris}
	\end{subfigure}
    \begin{subfigure}{.32\linewidth}
		\includegraphics[width=\linewidth]{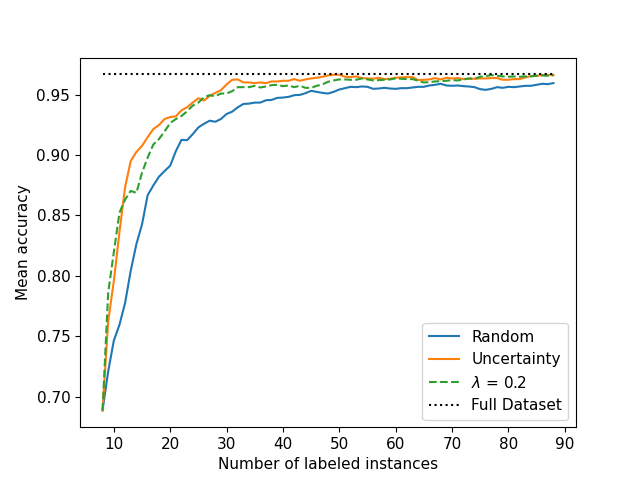}
		\caption{Wine}
	\end{subfigure}
    \caption{Mean accuracy vs. number of labeled instances for Random Sampling, Uncertainty Sampling, and the proposed method with $\lambda = 0.2$ for Iris and Wine.}\label{exp:al2}
\end{figure}

Even on datasets where the performance of the proposed method is lower, the gap is not always wide. Figure~\ref{exp:al2} shows two such datasets where the method performs less well. The difference is slightly greater between the proposed method and uncertainty sampling on the Iris dataset.

Since the objective here is to minimize labeling costs\footnote{Although it can also be to maximize performance given a cost}, one can set a performance threshold and focus on the actual reduction in costs. By accepting a 2\% loss in performance (threshold of 98\% performance on the full dataset), conventional uncertainty sampling reduces the number of labels in the Dog-2, Ionosphere, and Heart datasets by 76\%, 86\%, and 43\% respectively. In contrast, the proposed method reduces the number of labels in the respective datasets by 88\%, 91\%, and 83\%. However, a performance threshold has been set, which is why the area under the curve is a good indicator, as it captures the reduction in costs for all possible thresholds.
Table~\ref{exp:auac} shows the mean areas under the curve for the three methods studied and for each dataset. A statistical t-test is also performed between the first and second best methods for each values. Random sampling performs best on the Liver dataset, uncertainty sampling performs best on Seeds, Iris and Wine and the proposed method performs best on the other 10 datasets, except Blod where there is a tie.

\begin{table}[h]
    \caption{Mean AUAC for random sampling, uncertainty sampling and the proposed method with $\lambda = 0.2$ on each dataset. Student's t-test is also performed to determine the significance of the best method.}\label{exp:auac}%
    \begin{tabular*}{\textwidth}{@{\extracolsep\fill}lccccc@{}}
    \toprule
    & \multicolumn{3}{@{}c@{}}{Method} & \multicolumn{2}{@{}c@{}}{t-test} \\
    \cmidrule{2-4}\cmidrule{5-6}%
    Dataset & Random & Uncertainty & $\lambda = 0.2$ & statistic & p-value\\
    \midrule
    Bank & 81.17 & 81.49 & \textbf{82.23} & 2.74 & 0.0067\\
    Qsar & 97.69 & 99.15 & \textbf{99.16} & 0.35 & 0.7271\\
    Blod & 76.09 & \textbf{76.85} & \textbf{76.85} & 0.00 & 0.9965\\
    Breast Cancer & 93.87 & 94.96 & \textbf{95.31} & 1.84 & 0.0669\\
    Ionosphere & 75.77 & 81.06 & \textbf{82.40} & 2.33 & 0.0210\\
    Heart & 67.89 & 67.72 & \textbf{68.08} & 0.29 & 0.7741\\
    Liver & \textbf{58.07} & 57.37 & 58.02 & 0.07 & 0.9415\\
    Sonar & 67.79 & 70.67 & \textbf{70.94} & 0.37 & 0.7089\\
    Parkinson & 80.60 & 83.93 & \textbf{84.85} & 1.64 & 0.1034\\
    Dog-2 & 90.94 & 93.06 & \textbf{94.10} & 3.24 & 0.0014\\
    Seeds & 88.70 & \textbf{89.93} & 89.48 & 0.84 & 0.4010\\
    Iris & 88.22 & \textbf{91.23} & 90.60 & 1.19 & 0.2373\\
    Wine & 91.66 & \textbf{93.55} & 93.27 & 0.86 & 0.3920\\
    Glass & 57.33 & 58.32 & \textbf{59.05} & 0.87 & 0.3829\\
    Ecoli & 78.59 & 80.89 & \textbf{81.98} & 2.19 & 0.0300\\
    \botrule
    \end{tabular*}
\end{table}

\begin{figure}[h]%
    \centering
    \begin{subfigure}{.49\textwidth}
		\includegraphics[width=\textwidth]{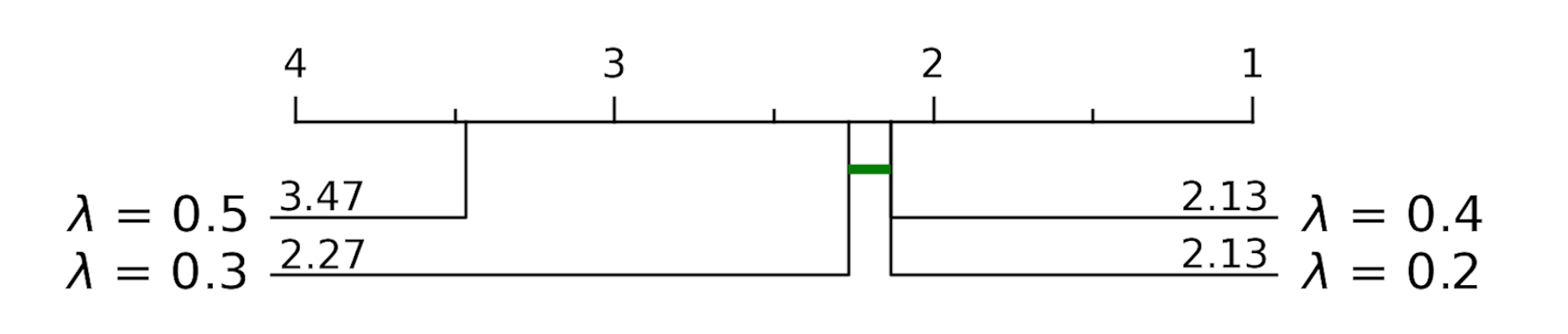}
		\caption{}\label{exp:cd1}
	\end{subfigure}
	\hfill
    \begin{subfigure}{.49\textwidth}
		\includegraphics[width=\textwidth]{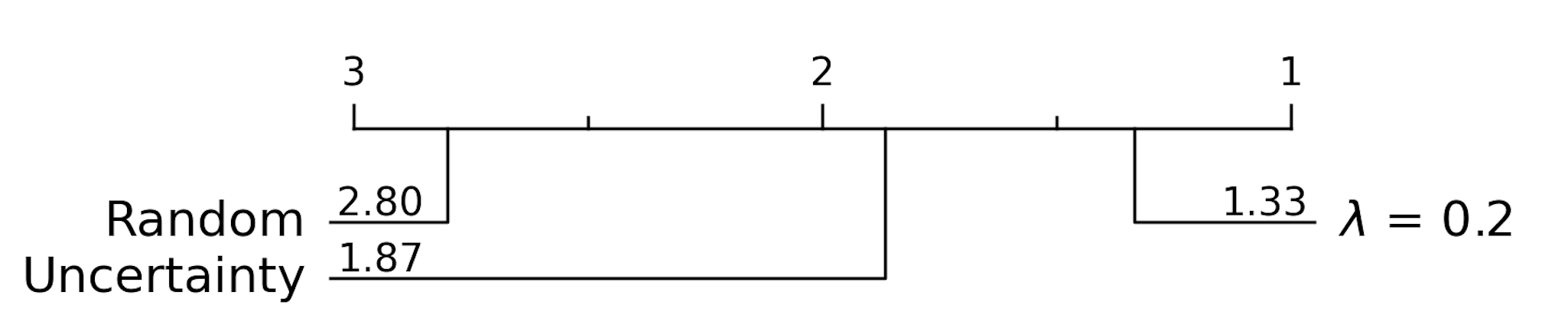}
		\caption{}\label{exp:cd2}
	\end{subfigure}
    \caption{Critical difference diagrams for different values of $\lambda$ on the proposed method~\ref{exp:cd1} and for random sampling, uncertainty sampling and the proposed method~\ref{exp:cd2}.}
\end{figure}

To statistically find the best model, a critical difference diagram is drawn up. The first diagram in Figure~\ref{exp:cd1} is a comparison of the proposed method with different values of $\lambda$. On all datasets, $\lambda = 0.2$ ranks on average at position $2.13$ out of $4$ and $\lambda = 0.5$ (which is equivalent to as much exploration as exploitation) ranks on average at position $3.47$. If a line connects two methods, this means that despite the better performance of one, the methods are not statistically differentiable. In the example, $\lambda = 0.2$, $\lambda = 0.3$ and $\lambda = 0.4$ are not statistically different.
Now, Figure~\ref{exp:cd2} is obtained by comparing the proposed method with random sampling and sampling by uncertainty. In average, the proposed method ranks $1.33$ out of $3$ and the significance of this result is demonstrated by the absence of a line linking the methods. It may also be interesting to note that uncertainty sampling and random sampling are not connected by a line either.

\section{Discussion} \label{section:discussion}

Calculating epistemic uncertainty (non-evidential) is demanding and not always accessible. It depends on the observations, requiring several phases of computation, including likelihood estimation, maximum likelihood, and optimization.

The two proposed methods offer simplicity, but there is a counterpart: the model must be capable of delivering a mass function to represent uncertainty and imprecision in the output. Such models exist but are not abundant. Among them are Evidential $K$-Nearest Neighbors~\citep{denoeux1995}, Evidential Decision Trees~\citep{ELOUEDI2001, Denoeux2000}, Evidential Random Forests~\citep{Hoarau2023}, and even Evidential Neural Networks~\citep{Yuan2020}. The proposed methods are compatible with probabilistic models (since a probability is a special belief function), but they may not capture the full depth of evidence modeling.

In the experiments above, $\lambda$ was set at $0.2$, meaning that the model prioritizes exploitation over exploration. This is a value that gives good results. Our studies to determine when one is more relevant than the other are illustrated in Figure~\ref{exp:cd1}. The results indicate that several lambda values yield fairly similar outcomes. The value of 0.2 is the one that gives the best performance in general: for the majority of datasets, it is more interesting to do more exploitation, without exceeding a certain limit, otherwise model performance will drop. For future work, it would be interesting to modify the value of $\lambda$ as the labeling process progresses. This adjustment could lead to a more powerful model capable of dynamically balancing between exploration and exploitation.

\section{Conclusion} \label{section:conclusion}

This paper introduces two new uncertainty sampling strategies and a novel representation method for them. These two methods use Klir uncertainty and an extended evidential epistemic uncertainty. A straightforward calculation on the model output enables the extraction of uncertainties. The objective is to also take into account the uncertainty present in richer labels, which was not possible up to now. The first strategy is based on Klir's uncertainty, combining discord (how self-conflicting the information is) and non-specificity (how imprecise the information is) in the model output. The second strategy extends epistemic (reducible) uncertainty to the evidential framework and to several classes, simplifying the computational phase.

The proposed Klir uncertainty sampling is chosen for its competitiveness in active learning. Its superiority over uncertainty sampling is statistically validated across the datasets examined in the experiments.
The novelty of this work lies in representing new information for uncertainty sampling, which also yields significant performance improvements in traditional active learning. The next step is to control exploration and exploitation (represented here as the $\lambda$ parameter) and to determine, for each dataset, when exploration or exploitation is more advantageous. The ability of the model to define areas of uncertainty, and to categorize these uncertainties, is then relevant information. 

\section*{Declarations}

\subsection*{Funding}
This work is funded by the Brittany region and the Côtes-d'Armor department. The authors also received funding from IRISA, the University of Rennes, DRUID and Orange SA.

\subsection*{Competing interests}
Arthur Hoarau, Jean-Christophe Dubois, Yolande Le Gall and Arnaud Martin received research support from the University of Rennes, the IRISA laboratory and the DRUID team. Vincent Lemaire received research support from Orange SA.

\subsection*{Ethics approval}
Not Applicable

\subsection*{Consent to participate}
All authors have read and approved the final manuscript.

\subsection*{Consent for publication}
All authors approved the publication.

\subsection*{Availability of data and materials}
All data used is this study are available publicly online. The datasets were extracted directly in the repositories available with the links in the folloing section.

\subsection*{Code availability}
The code for theoretical experiments is available at this link: \url{https://anonymous.4open.science/r/evidential-uncertainty-sampling-D453}. Link to the code for the experimental part in active learning: \url{https://anonymous.4open.science/r/evidential-active-learning-B266}.

\subsection*{Authors' contributions}
Arthur Hoarau, Vincent Lemaire, Jean-Christophe Dubois, Yolande Le Gall and Arnaud Martin contributed to the manuscript equally.

\bibliography{ref/main}

\end{document}